\documentclass[letterpaper]{article} 
\usepackage{aaai24}  
\usepackage{times}  
\usepackage{helvet}  
\usepackage{courier}  
\usepackage[hyphens]{url}  
\usepackage{graphicx} 
\usepackage{natbib}  
\usepackage{caption} 
\usepackage{algorithm}
\usepackage{algorithmic}

\usepackage{booktabs}       
\usepackage{amsmath,amssymb}
\usepackage{makecell}
\usepackage{colortbl}
\usepackage{hhline}
\usepackage{xcolor}
\usepackage{multirow}
\usepackage{subfigure}

\usepackage{newfloat}
\usepackage{listings}
\DeclareCaptionStyle{ruled}{labelfont=normalfont,labelsep=colon,strut=off} 
\floatstyle{ruled}
\newfloat{listing}{tb}{lst}{}
\floatname{listing}{Listing}
\title{Hyperbolic Graph Diffusion Model}
\author{
    Lingfeng Wen\textsuperscript{\rm 1},
    Xuan Tang\textsuperscript{\rm 2},
    Mingjie Ouyang\textsuperscript{\rm 1},\\
    Xiangxiang Shen\textsuperscript{\rm 1},
    Jian Yang\textsuperscript{\rm 3},
    Daxin Zhu\textsuperscript{\rm 4},
    Mingsong Chen\textsuperscript{\rm 1},
    Xian Wei\textsuperscript{\rm 1}\thanks{Corresponding author}
}
\affiliations{
    
    \textsuperscript{\rm 1}MoE Engineering Research Center of Hardware/Software Co-design Technology and Application, \\East China Normal University\\
    \textsuperscript{\rm 2}School of Communication and Electronic Engineering, East China Normal University\\

    \textsuperscript{\rm 3}Information Engineering University;
    \textsuperscript{\rm 4}Quanzhou Normal University\\
    xwei@sei.ecnu.edu.cn
}

\usepackage{bibentry}

\begin{document}

\maketitle

\begin{abstract}
Diffusion generative models (DMs) have achieved promising results in image and graph generation. However, real-world graphs, such as social networks, molecular graphs, and traffic graphs, generally share non-Euclidean topologies and hidden hierarchies. For example, the degree distributions of graphs are mostly power-law distributions. 
The current latent diffusion model embeds the hierarchical data in a Euclidean space, which leads to distortions and interferes with modeling the distribution. 
Instead, hyperbolic space has been found to be more suitable for capturing complex hierarchical structures due to its exponential growth property. 
In order to simultaneously utilize the data generation capabilities of diffusion models and the ability of hyperbolic embeddings to extract latent hierarchical distributions, we propose a novel graph generation method called, Hyperbolic Graph Diffusion Model (HGDM), which 
consists of an auto-encoder to encode nodes into successive hyperbolic embeddings, and a DM that operates in the hyperbolic latent space. 
HGDM captures the crucial graph structure distributions by constructing a hyperbolic potential node space that incorporates edge information. 
Extensive experiments show that HGDM achieves better performance in generic graph and molecule generation benchmarks, with a $48\%$ improvement in the quality of graph generation with highly hierarchical structures. 
\end{abstract}

\section{Introduction}
\begin{figure}[tbp]
\centering
    \includegraphics[width=0.9\columnwidth]{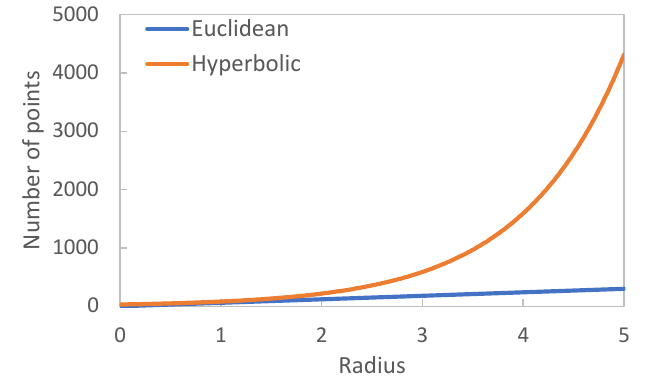}
    \caption{The number of points in 2D hyperbolic and Euclidean space that can be placed at radius $r$ from the center point while maintaining a distance of at least $s = 0.1$ from each other.}
    \label{capacity}
\end{figure}

Graph representation learning and graph generation are important machine learning tasks that have wide applications in fields such as social networks, drug discovery, and recommendation systems. While real-world graphs, such as social networks, molecular graphs, user-item interactions, and traffic graphs, generally have non-Euclidean topologies and hidden hierarchies \cite{asif2021graph,yang2017neural,fan2019graph}, e.g., the degree distribution of the nodes are mostly power-law distributions \cite{adamic2001search}. 
It is common practice in graph representation learning to embed node representations into Euclidean space \cite{chen2020graph}. 
However, the polynomially growing capacity of Euclidean space struggles to maintain the intrinsic distances of the huge number of nodes located in long-tailed regions and differentiate between them. Therefore, embedding these graphs into Euclidean space may produce distortions \cite{linial1995geometry}. 
Especially for the graphs of drug molecules, changes in a few atoms or chemical bonds can have a significant effect on the properties of the molecules \cite{thornber1979isosterism}.
On the contrary, it has been found that the capacity of hyperbolic spaces grows exponentially \cite{munzner1997h3} (see Figure \ref{capacity}). This property makes hyperbolic spaces well-suited for embedding tree-like or scale-free graphs \cite{chen2022modeling}. It has been observed that the complex hierarchical structures in hyperbolic embedding space become more prominent and easier to be captured \cite{mathieu2019continuous,nickel2018learning}. Some existing hyperbolic graph neural networks \cite{wu2021hyperbolic,liu2019hyperbolic} show excellent performance in graph representation learning tasks such as node classification and edge prediction. \citet{yang2022hicf,yang2022hrcf} also demonstrate superior performance over Euclidean methods in predicting long-tail items of user-item interaction for collaborative filtering in recommendation systems.

Furthermore, regarding generative models, diffusion models have achieved amazing results in synthesizing images \cite{rombach2022high}. 
The diffusion model perturbs the data distribution gradually during the forward diffusion process and then learns to recover the data distribution from noise through the reverse diffusion process. 
The Latent Diffusion Model \cite{rombach2022high} embeds images into a low-dimensional latent space to reduce the computational complexity. 
\citet{song2020score} unified \emph{Score matching with Langevin dynamics} (SMLD) \cite{song2019generative} and \emph{Denoising diffusion probabilistic modeling} (DDPM) \cite{ho2020denoising} into a system of stochastic differential equations (SDEs), referred as score-based generative models, which model the gradient of the log probability density of the data, the so-called ``score''. GDSS \cite{jo2022score} extended score-based generative models to graph generation tasks and proved that score-based models also outperform existing graph generative methods. However, this method performs in Euclidean space, which is inconsistent with the natural properties of graphs, does not adequately model the distribution of graph data, and still has room for improvement. In recent years, efforts have been made to extend the diffusion model to different manifolds to take full advantage of the underlying structure of the data. 
Torsional diffusion \cite{jing2022torsional} performs better for generating molecular conformers by applying diffusion processes on the hypertorus to the torsion angles. Riemannian diffusion models such as RDM \cite{huang2022riemannian} and RSGM \cite{de2022riemannian} extended the diffusion model to arbitrary compact and non-compact Riemannian manifolds.
However, they only test a few simple predefined distributions on manifolds and do not adequately evaluate the Riemannian diffusion model on larger real-world datasets.

In order to simultaneously utilize the excellent data generation capability of the diffusion model and the ability of the hyperbolic space to embed hierarchical distributions, we propose a two-stage graph diffusion model based on the hyperbolic latent space. 
Firstly, we train a hyperbolic graph variational auto-encoder (HVAE) to learn the hyperbolic representation of nodes. 
Then, we train two score models simultaneously to learn the reverse diffusion process in hyperbolic space (for nodes) and Euclidean space (for adjacency matrices), which also model the dependency between nodes and adjacency matrices. 
In addition, since diffusion models are more time-consuming for sampling compared to VAEs or GANs, in order to avoid another increase in time-consumption by using hyperbolic methods, we also design a \emph{Hyperbolic Graph Attention (HGAT)} layer as a basic module for HVAE and score-based models to maintain the performance without significantly increasing in time spent sampling. Our approach effectively utilizes the properties of hyperbolic space and the generative quality advantages of the diffusion model.

Our main contributions are summarized as follows:
\begin{itemize}
\item  We propose a novel approach for graph generation, the Hyperbolic Graph Diffusion Model (HGDM). To the best of our knowledge, 
HGDM is the first hyperbolic diffusion model in graph generation.
\item  We further design a simple and novel hyperbolic graph attention layer that utilizes the better representational capabilities of hyperbolic space without significantly increasing the computational time.
\item  
We consider both generic graph and molecular graph generation tasks. Experiments show that HGDM significantly outperforms all baselines in most metrics.
\end{itemize}
\section{Related Work}
\subsection{Graph Generation}
\citet{simonovsky2018graphvae} proposed a method called GraphVAE for generating small graphs using a variational auto-encoder (VAE). 
MolGAN \cite{de2018molgan} combines generative adversarial networks (GANs) and reinforcement learning objectives to encourage the generation of molecules with specific desired chemical properties. \cite{shi2020graphaf,zang2020moflow} generate molecules in a flow-based fashion. Graphdf \cite{luo2021graphdf} is an autoregressive flow-based model that utilizes discrete latent variables. \citet{liu2021graphebm} proposed an energy-based model to generate molecular graphs.
\subsection{Hyperbolic Graph Neural Network}
Hyperbolic Graph Neural Networks (HGNNs) \cite{liu2019hyperbolic,chami2019hyperbolic,dai2021hyperbolic,chen2021fully} utilize the characteristics of hyperbolic space to better capture the hierarchical structure in graph data. The core idea is to embed node representations into hyperbolic space and use hyperbolic distance to measure their relationships, enabling better handling of graph data with a hierarchical structure.
In the research of hyperbolic graph neural networks, 
\citet{chami2019hyperbolic} introduced the Hyperbolic Graph Convolutional Network (HGCN), which achieves new state-of-the-art results in learning embeddings for real-world graphs with hierarchical and scale-free features. 
Through experiments with hyperbolic auto-encoders, \citet{park2021unsupervised} find that utilizing hyperbolic geometry improves the performance of unsupervised tasks on graphs, such as node clustering and link prediction.
\citet{mathieu2019continuous} extended VAE into the hidden space of a Poincaré ball manifold as P-VAE, which shows better generalization to unseen data and the ability to qualitatively and quantitatively recover hierarchical structures. 
\subsection{Diffusion Models}
After the denoising diffusion probabilistic model \cite{ho2020denoising} was proposed, a large number of related works were carried out to improve this type of model such as DDIM \cite{DDIM2020}, score-based diffusion \cite{song2020score}. 
Some works extended diffusion models to graph generation tasks.
\citet{jo2022score} proposed a novel score-based graph generation model, which can generate both node features and adjacency matrices via the system of SDEs by introducing the diffusion process of graphs.
\citet{GeoDiff2022} proposed a novel generative model named GEODIFF for molecular conformation prediction. 
In addition, some works have extended the diffusion model to non-Euclidean manifolds to adequately model the data distribution.
\citet{Torsional2022} proposed torsional diffusion, a novel diffusion framework that operates on the space of torsion angles via a diffusion process on the hypertorus and an extrinsic-to-intrinsic score model. 
RDM \cite{huang2022riemannian} and RSGM \cite{de2022riemannian} extended the diffusion model to arbitrary compact and non-compact Riemannian manifolds, but have not been fully evaluated on more real-world datasets.

\section{Preliminaries}
\subsection{Riemannian Manifold}
A Riemannian manifold $(\mathcal{M},g^\mathcal{M})$ is a smooth manifold $\mathcal{M}$ equipped with a Riemannian metric $g^\mathcal{M}$ on the tangent space $\mathcal{T}_{x} \mathcal{M}$ at every point $x\in\mathcal{M}$. 
$\mathcal{T}_{x} \mathcal{M}$ is the vector space formed by all tangent vectors at $x$, which is locally a first-order approximation of the hyperbolic manifold at $x$. 
The Riemannian manifold metric $g^\mathcal{M}$ assigns a positive definite inner product $\langle\cdot,\cdot\rangle: \mathcal{T}_{x} \mathcal{M} \times \mathcal{T}_{x} \mathcal{M} \rightarrow \mathbb{R}$ on the tangent space, which makes it possible to define several geometric properties. 
For a curve $\gamma:[a,b]\rightarrow \mathcal{M}$, the length of $\gamma$ is defined as $ L(\gamma)=\int_a^b \left\| \gamma'(t) \right \| _g\ \mathrm{d}t, $where $\left \| \cdot \right \|_g$ denotes the norm induced by the Riemannian metric $g^\mathcal{M}$, i.e., $ \left \| v \right \| _g=\sqrt{g(v,v)}\quad$for any $ v\in T_x\mathcal{M}$. 
A geodesic is the analogue of a straight line in Euclidean geometry, being the shortest distance between two points $x,y$ on manifold $\mathcal{M}$: $d_\mathcal{M}(x,y)=\inf L(\gamma)$ where $\gamma$ is a curve with $\gamma(a)=x,\gamma(b)=y$. 
Exponential and logarithmic mappings are often described as additions and subtractions on Riemannian manifolds. 
The exponential map at point $x \in \mathcal{M}$, denoted as $\exp_x: T_x\mathcal{M}\rightarrow \mathcal{M}$, takes a tangent vector $v\in T_x\mathcal{M}$ to the point $y\in \mathcal{M}$ obtained by ``following'' the geodesic starting at $x$ in the direction of $v$ for a unit amount of time. 
The logarithmic map is the inverse of the exponential map $\log_x=\exp_x^{-1}: \mathcal{M} \rightarrow T_x\mathcal{M}$. 

\paragraph{Poincaré Ball Model.}
A hyperbolic manifold is a Riemannian manifold with a constant negative curvature c $(c<0)$. 
There exist multiple isomorphic hyperbolic models, including the Poincaré ball model, Lorentz model and Klein model. The Poincaré ball model is denoted as $(\mathcal{B}^n_c,g^\mathcal{B}_x)$, where $\mathcal{B}^n_c=\{x\in\mathbb{R}^n:\left \| x \right \|^2<-1/c\}$ is an open $n$-dimensional ball with radius $1/\sqrt{|c|}$. Its metric tensor is $g^\mathcal{B}_x=(\lambda^c_x)^2g^E_x$, with conformal factor $\lambda^c_x=2/(1+c\left \| x \right \|^2)$ and Euclidean metric $g^E$, i.e., $\mathbf{I}_n$. 
Given 2 points $\boldsymbol{x}, \boldsymbol{y}\in \mathcal{B}^n_c$, the induced geodesic distance on Poincaré ball is
\begin{equation}
\begin{scriptsize}
d_{\mathcal{B}}^{c}(\boldsymbol{x},\!\boldsymbol{y})\!=\!
\frac{1}{\sqrt{|c|}}\!\cosh ^{-1}\!\left(\!1\!-\!\frac{2 c\|\boldsymbol{x}-\boldsymbol{y}\|^{2}}{\left(1\!+\!c\|\boldsymbol{x}\|^{2}\right)\left(1\!+\!c\|\boldsymbol{y}\|^{2}\right)}\!\right)
\end{scriptsize}\!
\end{equation}
\citet{ganea2018hyperbolic} derived closed-form formulations for the exponential map as
\begin{equation}\label{2}
\exp _{\boldsymbol{x}}^{c}(\boldsymbol{v})=\boldsymbol{x} \oplus_{c}\left(\tanh \left(\sqrt{|c|} \frac{\lambda_{\boldsymbol{x}}^{c}\|\boldsymbol{v}\|}{2}\right) \frac{\boldsymbol{v}}{\sqrt{|c|}\|\boldsymbol{v}\|}\right)
\end{equation}
where $v\in T_x\mathcal{B}$. The logarithm map is given by
\begin{equation}\label{3}
\begin{scriptsize}
\log _{\boldsymbol{x}}^{c}(\boldsymbol{y})\!=\!\frac{2}{\sqrt{\!|c|} \lambda_{\boldsymbol{x}}^{c}}\! \tanh ^{\!-1}\!\left(\!\sqrt{\!|c|}\!\left\|-\boldsymbol{x}\! \oplus_{c}\! \boldsymbol{y}\right\|\!\right)\! \frac{-\boldsymbol{x} \oplus_{c} \boldsymbol{y}}{\left\|-\boldsymbol{x} \oplus_{c} \boldsymbol{y}\right\|}
\end{scriptsize}
\end{equation}
The parallel transport $PT_{x\rightarrow y}:T_x\mathcal{B} \rightarrow T_y\mathcal{B}$
defines the movement of a vector from one tangent space to another along a curve without changing itself, and is given by
\begin{equation}\label{4}
P T_{\mathbf{x} \rightarrow \mathbf{y}}^{c}(\mathbf{v})=\frac{\lambda_{x}^{c}}{\lambda_{\mathbf{y}}^{c}} \operatorname{gyr}[\mathbf{y},-\mathbf{x}] \mathbf{v}
\end{equation}
where $\oplus_{c}$ and $\operatorname{gyr}[\cdot,\cdot]v$ are Möbius addition \cite{ungar2007hyperbolic} and gyration operator \cite{ungar2008gyrovector}, respectively.
\subsection{Graph Diffusion Model}
Diffusion models perturb the data by adding progressively larger noise and then learn to reverse it. 
Recently, SMLD \cite{song2019generative} and DDPM \cite{ho2020denoising} are unified into the SDE system \cite{song2020score}. 
The Diffusion model perturbs the data in the forward SDE and estimates the score function (that is, the gradient of the log probability density with respect to data). Sampling is performed by using the estimated score function in the reverse SDE. 
GDSS \cite{jo2022score} extended the diffusion model to graph generation tasks. 
Graph diffusion models perturb both node features and adjacency matrix, modeling their dependencies to learn how to convert from noise to graph.

A graph $\boldsymbol{G}$ with $N$ nodes is defined by its node features $\boldsymbol{X} \in \mathbb{R}^{N\times F}$ and the weighted adjacency matrix $\boldsymbol{A} \in \mathbb{R}^{N\times N} $as $\boldsymbol{G} =(\boldsymbol{X},\boldsymbol{A}) \in \mathbb{R}^{N\times F}\times \mathbb{R}^{N\times N}:=\mathcal{G}$, where $F$ is the dimension of the node features.
Formally, the diffusion process can be represented as the trajectory of random variables$ \{\boldsymbol{G}_t = (\boldsymbol{X}_t,\boldsymbol{A}_t)\}_{t\in [0,T]}$ in a fixed time horizon $[0,T]$, where $\boldsymbol{G}_0$ is a graph from the data distribution $p_{data}$. The SDE of diffusion process is given by: 
\begin{equation}
\mathrm{d} \boldsymbol{G}_{t}=\boldsymbol{f}_{t}\left(\boldsymbol{G}_{t}\right) \mathrm{d} t+\boldsymbol{g}_{t}\left(\boldsymbol{G}_{t}\right) \mathrm{d} \mathbf{w}, \quad \boldsymbol{G}_{0} \sim p_{d a t a}
\end{equation}
where $\boldsymbol{f}_t(\cdot): \mathcal{G} \rightarrow \mathcal{G}$ is the linear drift coefficient, $\boldsymbol{g}_{t}(\cdot): \mathcal{G} \rightarrow \mathcal{G} \times \mathcal{G} $ is the diffusion coefficient, and $\mathbf{w}$ is the standard Wiener process.

The reverse-time SDE is separated into two parts in \cite{jo2022score}:
\begin{equation}
\begin{scriptsize}{
\left\{\hspace{-0.2cm}\begin{array}{l}
\mathrm{d} \boldsymbol{X}_{t}=\left[\boldsymbol{f}_{1, t}\left(\boldsymbol{X}_{t}\right)-\boldsymbol{g}_{1, t}^{2} \nabla_{\boldsymbol{X}_{t}} \log p\left(\boldsymbol{X}_{t}, \boldsymbol{A}_{t}\right)\right] \mathrm{d}t+\boldsymbol{g}_{1, t} \mathrm{~d} \overline{\mathbf{w}}_{1} \\
\mathrm{~d} \boldsymbol{A}_{t}=\left[\boldsymbol{f}_{2, t}\left(\boldsymbol{A}_{t}\right)-\boldsymbol{g}_{2, t}^{2} \nabla_{\boldsymbol{A}_{t}} \log p\left(\boldsymbol{X}_{t}, \boldsymbol{A}_{t}\right)\right] \mathrm{d}t+\boldsymbol{g}_{2, t} \mathrm{~d} \overline{\mathbf{w}}_{2}
\end{array}\right.
}\end{scriptsize}
\end{equation}
where $\boldsymbol{f}_{1,t}$ and $\boldsymbol{f}_{2,t}$ are linear drift coefficients
satisfying $\boldsymbol{f}_{t}(\boldsymbol{X}, \boldsymbol{A})=\left(\boldsymbol{f}_{1, t}(\boldsymbol{X}), \boldsymbol{f}_{2, t}(\boldsymbol{A})\right)$,
$\boldsymbol{g}_{1,t}$ and $\boldsymbol{g}_{2,t}$ are scalar diffusion coefficients, and  $\overline{\mathbf{w}}_{1}$,  $\overline{\mathbf{w}}_{2}$ are reverse-time standard Wiener processes. The $\nabla_{\boldsymbol{X}_{t}} \log p\left(\boldsymbol{X}_{t}, \boldsymbol{A}_{t}\right)$ and $\nabla_{\boldsymbol{A}_{t}} \log p\left(\boldsymbol{X}_{t}, \boldsymbol{A}_{t}\right)$ are 
named as the partial score functions.

\section{Hyperbolic Graph Diffusion Model}
In this section, we introduce the proposed Hyperbolic Graph Diffusion Model (HGDM).
Firstly, we introduce the hyperbolic wrapped normal distribution (HWN) and its sampling procedures.
Secondly, we describe how to perturb data in the hyperbolic space and the training objective. 
Thirdly, we introduce a novel hyperbolic graph attention layer and the architecture of our model. 
Finally, we present the sampling methods for the inverse diffusion process in the hyperbolic space.


\subsection{Probability Distributions on Hyperbolic Space}\label{section:4.1}
As an alternative to the normal distribution in Euclidean space, the hyperbolic wrapped normal distribution  \cite{nagano2019wrapped,mathieu2019continuous} is easy to sample and has a well-defined density function. For the Poincaré ball manifold, its density function is given by: 
\begin{equation}
\begin{array}{l}
\mathcal{N}_{\mathcal{B}_{c}^{d}}^{\mathrm{W}}(\boldsymbol{z} \mid \boldsymbol{\mu}, \Sigma)=\frac{d \nu^{\mathrm{W}}(\boldsymbol{z} \mid \boldsymbol{\mu}, \Sigma)}{d \mathcal{M}(\boldsymbol{z})} \\
=\mathcal{N}\left(\lambda_{\boldsymbol{\mu}}^{c} \log _{\boldsymbol{\mu}}(\boldsymbol{z}) \mid \mathbf{0}, \Sigma\right)\left(\frac{\sqrt{c} d_{p}^{c}(\boldsymbol{\mu}, \boldsymbol{z})}{\sinh \left(\sqrt{c} d_{p}^{c}(\boldsymbol{\mu}, \boldsymbol{z})\right)}\right)^{d-1}
\end{array}
\end{equation}
and its log density function is
\begin{equation}\label{8}
\begin{scriptsize}
\log p(\boldsymbol{z})\!=\!\log p(\boldsymbol{v})\!+\!(d-1)\log\! \left(\!\frac{ \sqrt{c} d_{p}^{c}(\boldsymbol{\mu}, \boldsymbol{z})}{\sinh\left(\sqrt{c} d_{p}^{c}(\boldsymbol{\mu}, \boldsymbol{z})\right)}\!\right)
\end{scriptsize}
\end{equation}
where $\boldsymbol{z}\in \mathcal{B}_{c}^{d}$, 
$p(\boldsymbol{v})$ is the normal distribution in the tangent space of origin $\mathcal{T}_{o} \mathcal{B}$.
 

This distribution can be constructed by defining Gaussian distribution on the tangent space at the origin of the hyperbolic space and projecting the distribution onto hyperbolic space after transporting the tangent space to a desired location in the space, using Eq. \eqref{4} and Eq. \eqref{2}. We use the HWN as the prior distribution on the hyperbolic manifold and it is convenient to compute gradient using Eq. \eqref{8}.

%
\subsection{Perturbing Data on Hyperbolic Space}
We consider two types of noise perturbations, the Variance Exploding (VE) and the Variance Preserving (VP) given in \cite{song2020score}. For Euclidean features, we have:  
\begin{equation}\label{9}
\begin{scriptsize}{
\begin{aligned}
&p(\mathbf{x}(t) \mid \mathbf{x}(0))=\\ &\left\{\begin{array}{ll}
\mathcal{N}\left(\mathbf{x}(0),\left[\sigma^{2}(t)-\sigma^{2}(0)\right] \mathbf{I}\right), & \text {(VE SDE) } \\
\mathcal{N}\left(\mathbf{x}(0)e^{-\frac{1}{2} \int_{0}^{t} \beta(s) \mathrm{d} s} , \mathbf{I}-\mathbf{I} e^{-\int_{0}^{t} \beta(s) \mathrm{d} s}\right), & (\text{VP SDE)} \\
\end{array}\right.
\end{aligned}
}\end{scriptsize}
\end{equation}
where $\sigma(t)=\sigma_{\min }\left(\frac{\sigma_{\max }}{\sigma_{\min }}\right)^{t} \text { for } t \in(0,1]$ and $\beta(t)=\bar{\beta}_{\min }+t\left(\bar{\beta}_{\max }-\bar{\beta}_{\min }\right) \text { for } t \in[0,1]$.

In order to parallelly train on different time steps, we directly give the hyperbolic distribution $p (x_t | x_0)$: 
\begin{equation}\label{10}
\begin{scriptsize}{
\begin{aligned}
    &p(\mathbf{x}(t) \mid \mathbf{x}(0))=\\ &\left\{\hspace{-0.1cm}\begin{array}{ll}
\mathcal{N}_{\mathcal{B}_{c}^{d}}^{\mathrm{W}}\!\left(\mathbf{x}(0),\left[\sigma^{2}(t)-\sigma^{2}(0)\right] \mathbf{I}\right), &\hspace{-0.25cm} \text{(VE SDE)} \\
\mathcal{N}_{\mathcal{B}_{c}^{d}}^{\mathrm{W}}\!\left(e^{-\frac{1}{2} \int_{0}^{t} \beta(s) \mathrm{d} s}\otimes_c \mathbf{x}(0) , \mathbf{I}-\mathbf{I} e^{-\int_{0}^{t} \beta(s) \mathrm{d} s}\right), &\hspace{-0.25cm} (\text{VP SDE)} \\
\end{array}\right.
\end{aligned}
}\end{scriptsize}
\end{equation}
where $\otimes_c$ is the Möbius scalar multiplication \cite{ungar2007hyperbolic}. 

\subsection{Training Objective}
The training process consists of two stages. Firstly, we train a hyperbolic graph variational auto-encoder (HVAE) to generate the hyperbolic representation of nodes. 
%
\begin{figure}[tbp]
\centering
    \includegraphics[width=0.9\columnwidth]{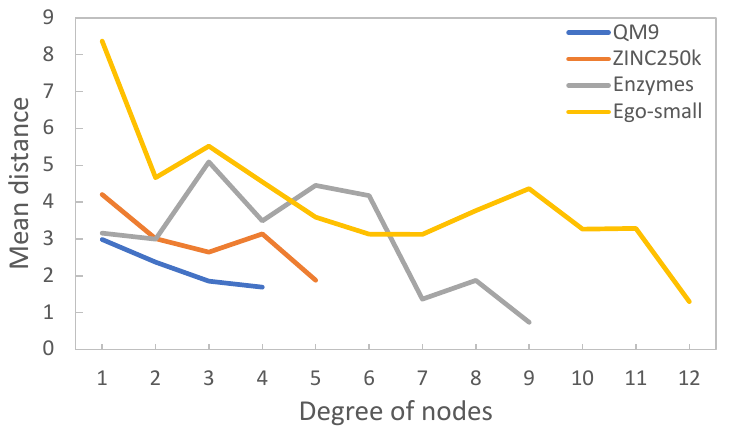}
    \caption{The average distance of the hyperbolic embedding of the nodes from the origin in different datasets with the degree of the nodes.}
    \label{distance}
\end{figure}
The training objective of the HVAE is as follows:
\begin{equation}
\begin{split}
\mathcal{L}_{\textit{VAE}}&=\mathcal{L}_{\textit{rec}}+ \mathcal{L}_{\textit{kl}}+ \mathcal{L}_{\textit{edge}} 
\end{split}
\end{equation}
which is a reconstruction loss combined with two regularization terms. The Kullback-Leibler (KL) regularization prevents latent embeddings from arbitrarily high variance and empirically we find it helpful for learning potential inverse diffusion processes. We also add a link prediction regularization objective, to encourage embedding to preserve the graph structure and implicitly constrain nodes with high degrees to be close to the origin (see Figure \ref{distance}). We use the Fermi-Dirac decoder to compute probability scores for edges:
\begin{equation}
\begin{scriptsize}
p\left(\boldsymbol{A}_{i, j} \ne 0 \mid \boldsymbol{x}^\mathcal{B}_{i}, \boldsymbol{x}^\mathcal{B}_{j}\right)=\left[e^{\left(d_{\mathcal{B}}^{c}\left(\boldsymbol{x}^\mathcal{B}_{i}, \boldsymbol{x}^\mathcal{B}_{j}\right)^{2}-r\right) / t}+1\right]^{-1}
\end{scriptsize}
\end{equation}

Secondly, following \citet{jo2022score}, we train two neural networks $\boldsymbol{s}_{\theta, t}$ and $\boldsymbol{s}_{\phi, t}$ simultaneously to estimate the partial score functions. 
But differently, given graph $\boldsymbol{G} =(\boldsymbol{X},\boldsymbol{A})\in \mathbb{R}^{N\times F}\times \mathbb{R}^{N\times N}$ from data, 
we first use a hyperbolic encoder to generate the hyperbolic latent feature $\boldsymbol{X}^\mathcal{B}_0\in \mathcal{B}^F_c$ of nodes, which combine with original adjacency matrix $\boldsymbol{A}$ forms $\bar{\boldsymbol{G}}_0=(\boldsymbol{X}^\mathcal{B}_0,\boldsymbol{A}_0)$. 
Then, we sample $\boldsymbol{\bar{G}}_{t}=(\boldsymbol{X}^\mathcal{B}_t,\boldsymbol{A}_t)$ using Eq. \eqref{9} and Eq. \eqref{10}.

We use $\boldsymbol{s}_{\theta, t}$ and $\boldsymbol{s}_{\phi, t}$  to estimate the gradients of the joint log-density so that $\boldsymbol{s}_{\theta, t}(\boldsymbol{\bar{G}}_{t}) \approx \nabla_{\boldsymbol{X}^\mathcal{B}_{t}} \log p\left(\boldsymbol{X}^\mathcal{B}_{t} \mid \boldsymbol{X}^\mathcal{B}_{0}\right)$ and $\boldsymbol{s}_{\phi, t}(\boldsymbol{\bar{G}}_{t}) \approx \nabla_{\boldsymbol{A}_{t}} \log p\left(\boldsymbol{A}_{t} \mid \boldsymbol{A}_{0}\right)$
. It is worth noticing that $\nabla_{\boldsymbol{X}^\mathcal{B}_{t}} \log p\left(\boldsymbol{X}^\mathcal{B}_{t} \mid \boldsymbol{X}^\mathcal{B}_{0}\right)$ is on $T_{\boldsymbol{X}^\mathcal{B}_{t}}\mathcal{B}$.

The training objectives are given by
\begin{equation}
\begin{scriptsize}
\begin{array}{l}
\min_{\theta}\! \mathbb{E}_{t}\!
\left\{\!\lambda_{1}\!(t) \mathbb{E}_{\boldsymbol {\bar G_{0}}}\! \mathbb{E}_{\boldsymbol{\bar G}_{t}\! \mid \!\boldsymbol{\bar G}_{0}\!}\!
\left\|\!\boldsymbol{s}_{\theta, t}\!\left(\!\boldsymbol{\bar G}_{t}\!\right)\!
-\!\nabla_{\!\boldsymbol{X}\!^\mathcal{B}_{t}}\! \log p\!\left(\!\boldsymbol{X}\!^\mathcal{B}_{t}\!\mid\!\boldsymbol{X}\!^\mathcal{B}_{0}\!\right)\!\right\|_{2}^{2}
\!\right\} 
\\
\min_{\phi}\! \mathbb{E}_{t}\!
\left\{\!\lambda_{2}\!(t) \mathbb{E}_{\boldsymbol{\bar G}_{0}}
\!\mathbb{E}_{\boldsymbol{\bar G}_{t}\! \mid\! \boldsymbol{\bar G}_{0}\!}\!
\left\|\boldsymbol{s}_{\phi, t}\!\left(\boldsymbol{\bar G}_{t}\right)\!
-\!\nabla_{\!\boldsymbol{A}_{t}}\! \log p\!\left(\boldsymbol{A}_{t}\! \mid \!\boldsymbol{A}_{0}\right)\right\|_{2}^{2}
\right\}
\end{array}
\end{scriptsize}
\end{equation}
where $\lambda_{1}(t)$ and $\lambda_{2}(t)$ are positive weighting functions and $t$ is uniformly sampled from $[0,T]$. 
\subsection{Hyperbolic Graph Attention Layer}
We propose the \emph{Hyperbolic Graph Attention (HGAT)} layer as the basic building block of HVAE and the node score predicting model $\boldsymbol{s}_{\theta, t}$, which is a variant of GAT\footnote{https://github.com/gordicaleksa/pytorch-GAT} and has smaller computational complexity than HGCN \cite{chami2019hyperbolic} (see Table \ref{Ablation}). 
The input of our layer is a set of hyperbolic node feature $\mathbf{h}^\mathcal{B}=\{\mathbf{h}^\mathcal{B}_1,\mathbf{h}^\mathcal{B}_2,...,\mathbf{h}^\mathcal{B}_N\},\mathbf{h}^\mathcal{B}_i\in \mathcal{B}^d$, where $N$ is the number of nodes, and $d$ is the dimension of features.

We first map $\mathbf{h}^\mathcal{B}$ into the tangent space of origin $T_o\mathcal{M}$ using Eq. \eqref{3} that $\mathbf{h}^E=\log _{\boldsymbol{o}}^{c}(\mathbf{h}^\mathcal{B})$ to perform the following attention aggregation mechanism \cite{velivckovic2017graph}. 

Next, we employ multi-head attention which is similar with \cite{velivckovic2017graph}, but we add adjacent feature $A_{ij}$ when computing attention $e^k_{i j}$ coefficients where $j\in \mathcal{N}_i$ 
\begin{equation}
e^{k}_{i j}=\text{leakyReLU}\left(\mathbf{W}^{k}_0 [\mathbf{h}^{E}_{i}, \mathbf{h}^{E}_{j},\mathbf{A}_{ij}]\right)
\end{equation}

We normalize them across all choices of $j$ using the softmax function as follows:
\begin{equation}
\alpha^{k}_{i j}=\mathrm{softmax}_j(e^{k}_{i j})
\end{equation}

Then, we concatenate features from $K$ head after computing the linear combination of the features corresponding to the normalized attention coefficients.
\begin{equation}
\mathbf{h}_{i}^{\prime E}=\mathop{\parallel}_{k=1}^{K}\left(\sum_{j \in \mathcal{N}_{i}} \alpha_{i j}^{k} \mathbf{W}^{k}_1 \mathbf{h}^{E}_{j}\right)
\end{equation}

We map the output $\mathbf{h}_{i}^{\prime E}$ of attention mechanism to the input $\mathbf{h}^B_i$ by using an exponential function, which can be regarded as a residual connection \cite{he2016deep} on the hyperbolic manifold, and finally we apply a manifold-preserving activation function $\sigma$ (e.g., ReLU or leaky ReLU) \cite{liu2019hyperbolic}.
\begin{equation}
\mathbf{h}_{i,\text{out}}^{B}=\sigma(\exp _{\boldsymbol{\mathbf{h}_{i}^B}}^{c}(P T_{\mathbf{o} \rightarrow \mathbf{h_{i}^B}}^{c}(\mathbf{h_{i}}^{\prime E})))
\end{equation}

\subsection{Hyperbolic Score-based Model}
We implement a hyperbolic variational auto-encoder to generate the hyperbolic representation $\boldsymbol{X}^\mathcal{B}$ of nodes. 
\paragraph{Encoder.}The encoder first utilizes a hyperbolic embedding layer to convert input feature$\boldsymbol{X}^E$ of nodes into hyperbolic embeddings $\boldsymbol{X}^{0,\mathcal{B}}$. $(\boldsymbol{X}^{0,\mathcal{B}},\boldsymbol{A})$ are then put into a $n$-layer HGAT to learn the structural information of the graph. Then the output $\boldsymbol{X}^{n,\mathcal{B}}$ is mapped to Euclidean space using Eq. \eqref{3} and fed into an MLP to produce mean $\mu$ and distortion $\sigma$. 
We use Eq. \eqref{2} to map $\mu$ to Poincaré ball and obtain $\boldsymbol{X}^\mathcal{B}$. 

\paragraph{Decoder.}
Firstly, we obtain the hyperbolic node feature $\boldsymbol{X}^{\mathcal{B}}=\{\boldsymbol{x}_1,\boldsymbol{x}_2,...,\boldsymbol{x}_{|V|}\}$ where $\boldsymbol{x}_{i}\in \mathcal{B}$ via reparameterization. 
After passing $(\boldsymbol{X}^{\mathcal{B}},\boldsymbol{A})$ to $k$ layers of HGAT, we get the output $\boldsymbol{X}^{k,\mathcal{B}}$. Then, the decoder uses a centroid-distance layer \cite{liu2019hyperbolic} to convert $\boldsymbol{X}^{k,\mathcal{B}}$ into the Euclidean space.
The centroid-distance layer is equipped with a list of learnable centroids $\mathcal{C}=[\boldsymbol{c}_1,\boldsymbol{c}_2,...,\boldsymbol{c}_{|\mathcal{C}|}]$ with each $\boldsymbol{c}_{i}\in \mathcal{B}$. 
By computing the pairwise distance $\psi_{ij}$ between $c_i$ and $x_j$, we obtain the Euclidean features while utilizing the hyperbolic metric and use it to recover $\boldsymbol{X}^{E}$.

\paragraph{Score Model.} 
We implement the hyperbolic score-based model $\boldsymbol{s}_{\theta, t}(\boldsymbol{\bar{G}}_{t})$ to estimate $\nabla_{\boldsymbol{X}^\mathcal{B}_{t}} \log p\left(\boldsymbol{X}^\mathcal{B}_{t},\boldsymbol{A}_{t} \right)$. The computing procedures of $\boldsymbol{s}_{\theta, t}$ consists:

\begin{equation}
\begin{scriptsize}{
\begin{aligned}
&\boldsymbol{V}^\prime=\operatorname{\textbf{MLP}}\!\left(\!\left[\left\{\log _{\boldsymbol{o}}^{c}\left(\boldsymbol{\operatorname{HGAT}\left(\boldsymbol{H}^\mathcal{B}_{i}, \boldsymbol{A}_{t}\right)}\right)\right\}_{i=0}^{K}\right]\!\right) \\
&\boldsymbol{V} = \log^c_{\boldsymbol{X}^\mathcal{B}_{t}} \left(\exp^c_o 
\left( \boldsymbol{V^\prime}\right) \right) * \operatorname{\textbf{MLP}}\left(t,\lambda^c_{\boldsymbol{X}^\mathcal{B}_t}\right)
\end{aligned}
}\end{scriptsize}
\end{equation}
where $\boldsymbol{V}=\boldsymbol{s}_{\theta, t}(\boldsymbol{\bar{G}}_{t}) $ is in the tangent space of $\boldsymbol{X}^\mathcal{B}_{t}$, $\boldsymbol{H}^\mathcal{B}_{i+1} = \operatorname{HGAT}(\boldsymbol{H}^\mathcal{B}_i, A_t)$ with $\boldsymbol{H}^\mathcal{B}_0 = \boldsymbol{X}^\mathcal{B}_t$ being given, $[\cdot]$ denotes the concatenation operation, and $K$ denotes the number of HGAT layers.
Unlike the score in Euclidean space, we find that the norm of hyperbolic scores is related to the Poincaré metric $g^\mathcal{B}_x=(\lambda^c_x)^2g^E_x$, 
but the metric varies with $x$, which makes it difficult to match the score. We use an MLP to rescale the output according to the time step and the conformal factor $\lambda^c_x$, which empirically helps the model capture the norm of the true score.

We use the score-based model $\boldsymbol{s}_{\phi, t}(\boldsymbol{\bar{G}}_{t}) $ to estimate $\nabla_{\boldsymbol{A}_{t}} \log p\left(\boldsymbol{X}^\mathcal{B}_{t},\boldsymbol{A}_{t} \right)$ as follows:
\begin{equation}
\begin{scriptsize}
\boldsymbol{s}_{\phi, t}\left(\boldsymbol{\bar{G}}_{t}\right)=\operatorname{\textbf{MLP}}
\left(\left[\left\{\left(\boldsymbol{\boldsymbol{H}^E_{i}}\right)\right\}_{i=0}^{L}\right]\right)
\end{scriptsize}
\end{equation}
where $\boldsymbol{H}^E_{i+1} = \operatorname{\mathrm{GNN}}(\boldsymbol{H}^E_i,A_t)$ with $\boldsymbol{H}^E_0$ obtained by a centroid-distance layer and $L$ is the number of GNN layers. 
\subsection{Reverse Diffusion Sampling on Hyperbolic Space}
\begin{algorithm}[t]
\caption{Hyperbolic PC sampling (VE SDE)}
\label{alg:algorithm1}
\begin{algorithmic}[1] 
\STATE $x_N\sim \mathcal{N}_{\mathcal{B}_{c}^{d}}^{\mathrm{W}}(0,\sigma^2_{max} \mathbf{I})$
\FOR{$i = N  - 1 $ \textbf{to} $ 0$}
    \STATE $x^\prime_i \gets \exp _{\boldsymbol{x_{i+1}}}^{c}((\sigma^2_{i+1}-\sigma^2_i)\boldsymbol{s}_\theta(x_{i+1},A_{i+1}))$\\
    \STATE $x_i\sim \mathcal{N}_{\mathcal{B}_{c}^{d}}^{\mathrm{W}}(x^\prime_i,\sqrt{\sigma^2_{i+1}-\sigma^2_i} \mathbf{I})$\\
    \FOR{$j = 1 $ \textbf{to} $M$}
        \STATE $x^{\prime\prime}_i \gets  \exp _{\boldsymbol{x_{i}}}^{c}(\epsilon_i\boldsymbol{s}_\theta(x_{i},A_{i}))$ \\
        \STATE $x_i \sim \mathcal{N}_{\mathcal{B}_{c}^{d}}^{\mathrm{W}}(x^{\prime\prime}_i,\sqrt{2\epsilon_i} \mathbf{I}) $
    \ENDFOR
\ENDFOR
\STATE \textbf{return} $x_0$
\end{algorithmic}
\end{algorithm}
\begin{algorithm}[t]
\caption{Hyperbolic PC sampling (VP SDE)}
\label{alg:algorithm2}
\begin{algorithmic}[1] 
\STATE $x_N\sim \mathcal{N}_{\mathcal{B}_{c}^{d}}^{\mathrm{W}}(0, \mathbf{I})$
\FOR{$i = N  - 1 $ \textbf{to} $ 0$}
    \STATE $x^\prime_i \gets (2-\sqrt{1-\beta_{i+1}})\otimes_c\boldsymbol{x_{i+1}}$\\
    \STATE $x^\prime_i \gets \exp _{\boldsymbol{x^\prime_i}}^{c}(\beta_{i+1}\boldsymbol{s}_\theta(x_{i+1},A_{i+1}))$\\
    \STATE $x_i\sim \mathcal{N}_{\mathbb{B}_{c}^{d}}^{\mathrm{W}}(x^\prime_i,\sqrt{\beta_{i+1}} \mathbf{I})$\\
    \FOR{$j = 1 $ \textbf{to} $M$}
        \STATE $x^{\prime\prime}_i \gets  \exp_{\boldsymbol{x_{i}}}^{c}(\epsilon_i\boldsymbol{s}_\theta(x_{i},A_{i}))$ \\
        \STATE $x_i \sim \mathcal{N}_{\mathcal{B}_{c}^{d}}^{\mathrm{W}}(x^{\prime\prime}_i,\sqrt{2\epsilon_i} \mathbf{I}) $
\ENDFOR
\ENDFOR
\STATE \textbf{return} $x_0$
\end{algorithmic}
\end{algorithm}

\begin{table*}[ht]
  
  \Huge
  \centering

  \resizebox{\linewidth}{0.13\hsize}{
  \begin{tabular}{cccc>{\columncolor{gray!20}}cccc>{\columncolor{gray!20}}cccc>{\columncolor{gray!20}}cccc>{\columncolor{gray!20}}c}
    \toprule
    
    Dataset&\multicolumn{4}{c}{Ego-small}&\multicolumn{4}{c}{Community-small}&\multicolumn{4}{c}{Enzymes}&\multicolumn{4}{c}{Grid} \\
    Info.&\multicolumn{4}{c}{Real, $4 \le |V | \le 18$, $\delta=0.25$}&\multicolumn{4}{c}{Synthetic, $12 \le |V | \le 20$, $\delta=1$}&\multicolumn{4}{c}{Real, $10 \le |V | \le 125$, $\delta=1.26$}&\multicolumn{4}{c}{Synthetic, $100 \le |V | \le 400$ , $\delta=9.76$} \\
    \hhline{~*{16}{-}}
    Method&Deg.&Clus.&Orbit&Avg.&Deg.&Clus.&Orbit&Avg.&Deg.&Clus.&Orbit&Avg.&Deg.&Clus.&Orbit&Avg.\\
    \midrule
    DeepGMG&0.040&0.100&0.020&0.053&0.220&0.950&0.400&0.523&-&-&- &-&-&-&-&-\\
    GraphRNN&0.090&0.220&\underline{0.003}&0.104&0.080&0.120&0.040&0.080&\textbf{0.017}&0.062&0.046&\underline{0.042}&\textbf{0.064}&0.043&\textbf{0.021}&\textbf{0.043}\\
    GraphAF&0.03&0.11&\textbf{0.001}&0.047&0.18&0.20&0.02&0.133&1.669&1.283&0.266&1.073&-&-&-&-\\
    GraphDF&0.04&0.13&0.01&0.060&0.06&0.12&0.03&0.070&1.503&1.061&0.202&0.922&-&-&-&-\\
    GraphVAE&0.130&0.170&0.050&0.117&0.350&0.980&0.540&0.623&1.369&0.629&0.191&0.730&1.619&\textbf{0.0}&0.919&0.846\\
    GNF&0.030&0.100&\textbf{0.001}&0.044&0.200&0.200&0.110&0.170&-&-&-&-&-&-&-&-\\
    EDP-GNN&0.052&0.093&0.007&0.051&0.053&0.144&0.026&0.074&\underline{0.023}&0.268&0.082&0.124&0.455&0.238&0.328&0.340\\
    GDSS&\underline{0.021}&\underline{0.024}&0.007&\underline{0.017}&\underline{0.045}&\underline{0.086}&\underline{0.007}&\underline{0.046}&0.026&\underline{0.061}&\underline{0.009} &\textbf{0.032}&\underline{0.111}&0.005&0.070&\underline{0.062}\\
    \midrule
    HGDM (ours)&\textbf{0.015}&\textbf{0.023}&\underline{0.003}&\textbf{0.014}&\textbf{0.017}&\textbf{0.050}&\textbf{0.005}&\textbf{0.024}&0.045&\textbf{0.049}&\textbf{0.003} &\textbf{0.032}& 0.137&\underline{0.004} &\underline{0.048} & 0.063\\
    Improve. over GDSS &28.6\%&4.2\%&57.1\%&17.6\%& 62.2\%&41.9\%&28.6\%&47.8\%&-73.1\%&19.7\%&66.7\%&0.0\%&-23.4\%&20.0\%&31.4\%&-1.6\%\\
    \bottomrule
  \end{tabular}
  }
  \caption{Generation results on the generic graph datasets. Results of the baselines are taken from published papers \cite{niu2020permutation,luo2021graphdf,jo2022score}. Hyphen (-) denotes that the results are not provided in the original paper. The best results are highlighted in bold and the underline denotes the second best. (lower is better). We report the mean graph hyperbolicity values $\delta$ of all datasets. Due to the space limitation, we provide the standard deviations in Appendix.}
\label{generic}
\end{table*}
We implement the Predictor-Corrector samplers \cite{song2020score} in hyperbolic space (see Algorithms \ref{alg:algorithm1} and \ref{alg:algorithm2}). 
We use the Möbius scalar multiplication $\otimes_c$ \cite{ungar2007hyperbolic} and the exponential map in Eq. \eqref{2} to keep the reverse diffusion process on the manifold.
$\{\epsilon_i\}_{i=0}^{N-1}$ are step sizes for Langevin dynamics. 
After sampling, a pre-trained decoder from HVAE is used to recover $x_0$ back into the original node features, such as atom types.

\section{Experimental Results}

In this section, we evaluate our method on generic graph datasets and molecular datasets, then compare it with the baselines. 
\subsection{Generic Graph Generation}
\paragraph{Experimental Setup}
We validate HGDM on four generic graph datasets: (1) \textbf{Ego-small}, 200 small ego graphs drawn from larger Citeseer network dataset \cite{sen2008collective}, (2) \textbf{Community-small}, 100 randomly generated community graphs \cite{jo2022score}, (3) \textbf{Enzymes}, 587 protein graphs which represent the protein tertiary structures of the enzymes from the BRENDA database \cite{schomburg2004brenda}, and (4) \textbf{Grid}, 100 standard 2D grid graphs \cite{jo2022score}. We use the maximum mean discrepancy (MMD) to compare the distributions of graph statistics between the same number of generated and test graphs. Following \citet{jo2022score}, we measure the distributions of degree, clustering coefficient, and the number of occurrences of orbits with 4 nodes. We also report mean graph hyperbolicity values $\delta$ (lower is more hyperbolic) of the datasets by computing Gromovs $\delta$-hyperbolicity \cite{adcock2013tree}, a notion from group theory that measures how tree-like a graph is. The lower $\delta$, the more hyperbolic the graph dataset, and $\delta=0$ for trees.

\paragraph{Baselines}
We compare our proposed method against the following generative models. \textbf{GraphVAE} \cite{simonovsky2018graphvae} is a VAE-based model. \textbf{DeepGMG} \cite{li2018learning} and \textbf{GraphRNN} \cite{you2018graphrnn} are autoregressive RNN-based models. \textbf{GNF} \cite{liu2019graph} is a one-shot flow-based model. \textbf{GraphAF} \cite{shi2020graphaf} is an autoregressive flow-based model. \textbf{EDP-GNN} \cite{niu2020permutation} and \textbf{GDSS} \cite{jo2022score} are score-based models. \textbf{GraphDF} \cite{luo2021graphdf} is an autoregressive flow-based model that utilizes discrete latent variables. 
\paragraph{Results}
Table \ref{generic} shows that HGDM significantly outperforms all baselines including GDSS, achieving optimal results in most metrics. In particular, HGDM outperforms GDSS with a 37.1\% decrease of MMD on average in Ego-small and Community-small, indicating that it has unique advantages in generating graphs with low hyperbolicity $(\delta\le 1)$. 
Although the graphs in Enzymes deviate from a power-law distribution,
HGDM still maintains the same results as GDSS in terms of average statistics.
The graphs in Grid deviate more from the hierarchical structure $(\delta=9.76)$ and are closer to the Euclidean topology with 0 curvature, in which case the autoregressive model GraphRNN has an advantage over the score-based models such as GDSS. 
HGDM still manages to achieve comparable results to GDSS on average statistics and outperforms the other baselines except GraphRNN.
This demonstrates the scalability of HGDM, which still has good modeling capability for graphs with high hyperbolicity.
In general, we find that our model performs better in generating graphs with small $\delta$-hyperbolicity. 

\subsection{Molecule Generation}

\begin{table*}[th]

  \centering
  \resizebox{\linewidth}{!}{
  \begin{tabular}{ccccccccc}
    \toprule
    Dataset&\multicolumn{4}{c}{QM9}&\multicolumn{4}{c}{ZINC250k} \\
    Info.&\multicolumn{4}{c}{$\delta=0.7$}&\multicolumn{4}{c}{$\delta=1$} \\
    \cline{2-9}
     Method & \makecell{Val. w/o corr. (\%)} $\uparrow$ & \makecell{NSPDK MMD} $\downarrow$ & FCD $\downarrow$&time (s) $\downarrow$& \makecell{Val. w/o  corr. (\%) }$\uparrow$ & NSPDK MMD $\downarrow$ & FCD $\downarrow$ &time (s) $\downarrow$\\
    \midrule
    GraphAF & 67 & 0.020 & 5.268 &2.52$e^3$& 68 & 0.044 & \textbf{16.289}&5.80$e^3$ \\
    GraphDF & 82.67 & 0.063 & 10.816 &5.35$e^4$& 89.03 & 0.176 &34.202&6.02$e^3$  \\
    MoFlow & 91.36 & 0.017 & 4.467&\textbf{4.60}& 63.11&0.046 & 20.931&\textbf{2.45}$\mathbf{e^1}$\\
    EDP-GNN & 47.52 & 0.005 & 2.680&4.40$e^3$& 82.97 & 0.049&16.737&9.09$e^3$  \\
    GraphEBM & 8.22 & 0.030 & 6.143&\underline{3.71$e^1$}& 5.29 & 0.212 & 35.471&\underline{5.46$e^1$}  \\
    $\text{GDSS}^*$ & \underline{95.79}  & \underline{0.003} & \underline{2.813} &1.14$e^2$& \textbf{95.90} & \underline{0.019} & \underline{16.621}&2.02$e^3$ \\
    \midrule
    HGDM (ours)& \textbf{98.04} & \textbf{0.002} & \textbf{2.131} & 1.23$e^2$ & \underline{93.51} & \textbf{0.016} & {17.69} & 2.23$e^3$\\
    \bottomrule
  \end{tabular}
  }
  \caption{Generation results on the QM9 and ZINC250k dataset. Results of the baselines are taken from \cite{jo2022score}. The method with $*$ denotes that results are obtained by running open-source codes. The best results are highlighted in bold and the underline denotes the second best. We report the mean graph hyperbolicity values $\delta$ of the two datasets. Due to the space limitation, we show the results of validity, uniqueness, and novelty in the Appendix.}
\label{Molecule}
\end{table*}
\paragraph{Datasets}
Following the GDSS \cite{jo2022score}, we tested our model on the \textbf{QM9} \cite{ramakrishnan2014quantum} and \textbf{ZINC250k} \cite{irwin2012zinc} datasets to assess its ability to learn molecular distributions. 
%
QM9 dataset contains 134k stable small molecules with up to $9$ heavy atoms (CONF).
ZINC250k datasets encompass a total of 250k chemical compounds with drug-like properties. On average, these molecules are characterized by a larger size (with an average of 23 heavy atoms) and possess greater structural complexity when compared to the molecules in QM9. Following previous works \cite{luo2021graphdf,jo2022score}, the molecules are kekulized by the RDKit library \cite{landrum2016rdkit} with hydrogen atoms removed.
We also report the mean graph hyperbolicity values $\delta$.

\paragraph{Metrics}
 We sample $10,000$ molecules using our model and evaluate their quality with the following metrics. \textbf{Fréchet ChemNet Distance (FCD)} \cite{preuer2018frechet} evaluates the distance between the training sets and generated sets by analyzing the activations of the penultimate layer of the ChemNet. \textbf{Neighborhood Subgraph Pairwise Distance Kernel (NSPDK) MMD} \cite{costa2010fast} computes MMD between the generated molecules and test molecules while considering both node and edge features for evaluation. It is important to note that FCD and NSPDK MMD are crucial metrics that evaluate the ability to learn the distribution of the training molecules by measuring the proximity of the generated molecules to the distribution. Specifically, FCD assesses the ability in the context of chemical space, whereas NSPDK MMD measures the ability in terms of the graph structure. \textbf{Validity w/o correction},  used for fair comparing with \cite{jo2022score}, is the proportion of valid molecules without valency correction or edge resampling. \textbf{Time} measures the time for generating $10,000$ molecules in the form of RDKit molecules.

\paragraph{Baselines}
We use \textbf{GDSS} \cite{jo2022score} as our main baseline. We also compare the performance of HGDM against \textbf{GraphAF} \cite{shi2020graphaf}, \textbf{GraphDF} \cite{luo2021graphdf}, MoF\textbf{}low \cite{zang2020moflow}, \textbf{EDP-GNN} \cite{niu2020permutation} and \textbf{GraphEBM} \cite{liu2021graphebm}.
\begin{table}
\Huge
  \centering
  \resizebox{\linewidth}{0.13\hsize}{
  \begin{tabular}{ccccc}
    \toprule
        Method &\makecell{Val. w/o  corr. (\%) }$\uparrow$ & NSPDK MMD $\downarrow$ & FCD $\downarrow$&time (s) $\downarrow$\\
    \midrule
    GDSS&95.79& 0.003& 2.813&\textbf{1.14}$\mathbf{e^2}$ \\
    GDSS+AE & 95.95 & 0.003 & 2.615&1.17$e^2$ \\
    HGDM+hgcn(ours)& 96.64 & \textbf{0.002} & 2.331  &1.48$e^2$ \\
    HGDM (ours)& \textbf{98.04} & \textbf{0.002} & \textbf{2.131}& 1.23$e^2$  \\

    \bottomrule
  \end{tabular}
  }
  \caption{Generation results of the variants of GDSS and HGDM on the QM9 dataset.}
  \label{Ablation}
\end{table}

\paragraph{Results}
Table \ref{Molecule} shows that our method also exhibits excellent performance on molecular generation tasks. In particular, on the QM9 dataset, our method is optimal in all metrics and significantly outperforms our main comparator, GDSS. HGDM achieves the highest validity without the use of post-hoc valency correction. It shows that HGDM can effectively learn the valency rules of the molecules. HGDM also outperforms all baselines in NSPDK MMD and FCD, which indicates that not only does the HGDM efficiently learn the distribution of the graph structure from the hyperbolic latent space and benefit from it, but the generated molecules are also close to the data distribution in the chemical space.
We also report the generation results of ZINC250k in Table \ref{Molecule}. As we observed in the generic graph datasets, the performance of HGDM in FCD and validity is slightly degraded due to the fact that the ZINC250k dataset deviates more from the hyperbolic structure and has more complex chemical properties compared to QM9. However, HGDM still maintains the advantage of generation from a geometric structure perspective that outperforms all the baselines in NSPDK MMD. Overall HGDM achieves similar performance for GDSS on the ZINC250k dataset and outperforms all other baselines. The superior performance of HGDM on the molecule generation task validates the ability of our method to efficiently learn the underlying distribution of molecular graphs with multiple nodes and edge types.

\subsection{Ablation Study}

We implement a variant of GDSS called GDSS+AE. It incorporates an autoencoder to generate the Euclidean embedding of nodes and uses GDSS to estimate the score function in Euclidean latent space. We use GDSS+AE to compare the effect of Euclidean and hyperbolic hidden spaces on graph diffusion. In addition, we tested HGDM using the HGCN layer as a building block, called HGDM+hgcn. 
The results of the above-mentioned variants in QM9 are provided in Table \ref{Ablation}
\paragraph{Necessity of Hyperbolic Hidden Space}
We find that GDSS+AE is slightly improved in metrics compared to GDSS. 
It can be considered that the dense embedding generated by the auto-encoder helps to learn the distribution of the graph, but it is not comparable to our hyperbolic approach due to the restricted capacity growth of Euclidean space. 
\paragraph{Necessity of HGAT Layers}
HGDM+hgcn is comparable to our model in most metrics, but the increase in time cost is higher compared to GDSS, whereas our method has only a slight increase in time, making it more suitable for scaling up to the task of generating large graphs. 
This demonstrates the importance of using our proposed HGAT layer.
\section{Conclusion}

In this work, we proposed a two-stage Hyperbolic Graph Diffusion Model (HGDM) to learn better the distribution of graph data and a simple hyperbolic graph attention layer that reduces the extra time spent associated with hyperbolic methods.
HGDM overcomes the limitations of previous graph generation methods and is closer to the nature of graphs.
We have found experimentally that learning the distribution in the hyperbolic space is beneficial for the quality of generated graph with the power-law distribution. 
Experimental results in generic graph and molecule generation show that 
HGDM outperforms existing graph generation methods in most metrics, demonstrating the importance of learning the underlying manifold for graph generation tasks.
Code is available at https://github.com/LF-WEN/HGDM
\section*{Acknowledgments}
This work is supported by the National Natural Science Foundation of China (42130112,62272170). This work
is also supported by the ``Digital Silk Road'' Shanghai International Joint Lab of Trustworthy Intelligent Software (22510750100), the General Program of Shanghai Natural Science Foundation (23ZR1419300) and KartoBit Research Network(KRN2201CA).

\bigskip

\bibliography{aaai24}

\appendix

\setcounter{secnumdepth}{1}
\section{Additional Experimental Results}
In this section, we provide additional experimental results.

\subsection{Generic Graph Generation}
We report the standard deviation of the generation results of Table \ref{generic} in Table \ref{Additional0} and Table \ref{Additional1}.

\subsection{Molecule Generation}
We additionally report the validity, uniqueness, and novelty of the generated molecules as well as the standard deviation of the results in Table \ref{Additional2} and Table \ref{Additional3}. \textbf{Validity} is the fraction of the generated molecules that do not violate the chemical valency rule. \textbf{Uniqueness} is the fraction of the valid molecules that are unique. \textbf{Novelty} is the fraction of the valid molecules that are not included in the training set.

We observe that the molecules generated by HGDM have a lower novelty in Table \ref{Additional2}. This can be interpreted that high novelty may not represent good generation quality for the model in the QM9 dataset and some models with poorer generation quality such as GraphDF and GraphEBM still achieve high novelty. As discussed in \cite{vignac2021top,hoogeboom2022equivariant}, QM9 is an enumeration of all possible molecules up to 9 heavy atoms that satisfy a predefined set of constraints, a molecule that is novel does not satisfy at least one of these constraints, which means that the algorithm failed to capture some properties of the dataset.
\begin{table*}[htbp]
  \centering
  \resizebox{\linewidth}{!}{

  \begin{tabular}{ccccccc}
    \toprule
    Dataset&\multicolumn{3}{c}{Ego-small}&\multicolumn{3}{c}{Community-small}\\
    \cmidrule(r){2-4}\cmidrule(r){5-7}
    Method&Deg.&Clus.&Orbit&Deg.&Clus.&Orbit\\
    \midrule
    GDSS&0.021$\pm$0.008&0.024$\pm$0.007&0.007$\pm$0.005&0.045$\pm$0.028&0.086$\pm$0.022&0.007$\pm$0.004\\
    HGDM (ours)&\textbf{0.015}$\pm$0.005&\textbf{0.023}$\pm$0.006&\textbf{0.003}$\pm$0.005&\textbf{0.017}$\pm$0.029&\textbf{0.050}$\pm$0.018&\textbf{0.005}$\pm$0.003\\
    \bottomrule
  \end{tabular}
  }
  \caption{\textbf{Generation results of HGDM on the Ego-small and the Community-small datasets.} Results of GDSS are taken from published papers \cite{jo2022score}. The best results are highlighted in bold (lower is better). We report the MMD distance between the test datasets and generated graphs with the standard deviation.}
\label{Additional0}
\end{table*}

\begin{table*}[htbp]
  \centering
  \resizebox{\linewidth}{!}{
  \begin{tabular}{ccccccc}
    \toprule
    Dataset&\multicolumn{3}{c}{Enzymes}&\multicolumn{3}{c}{Grid} \\
    \cmidrule(r){2-4}\cmidrule(r){5-7}
    Method&Deg.&Clus.&Orbit&Deg.&Clus.&Orbit\\
    \midrule
    
    GraphRNN&\textbf{0.017}$\pm$ 0.007&0.062 $\pm$ 0.020&0.046 $\pm$ 0.031&\textbf{0.064} $\pm$ 0.017&0.043 $\pm$ 0.022&\textbf{0.021} $\pm$ 0.007\\
    GraphAF&1.669 $\pm$ 0.024&1.283 $\pm$ 0.019&0.266 $\pm$ 0.007&-&-&-\\
    GraphDF&1.503 $\pm$ 0.011&1.061 $\pm$ 0.011&0.202 $\pm$ 0.002&-&-&-\\
    GraphVAE&1.369 $\pm$ 0.020&0.629 $\pm$ 0.005&0.191 $\pm$ 0.020&1.619 $\pm$ 0.007&\textbf{0.0} $\pm$ 0.000&0.919 $\pm$ 0.002\\
    EDP-GNN&0.023 $\pm$ 0.012&0.268 $\pm$ 0.164&0.082 $\pm$ 0.078&0.455 $\pm$ 0.319&0.238 $\pm$ 0.380&0.328 $\pm$ 0.278\\
   GDSS&0.026 $\pm$ 0.008&0.061 $\pm$ 0.010&0.009 $\pm$ 0.005&0.111 $\pm$ 0.012&0.005 $\pm$ 0.000&0.070 $\pm$ 0.044\\
    HGDM (ours)&0.045$\pm$0.008&\textbf{0.049}$\pm$0.011&\textbf{0.003}$\pm$0.001& 0.137$\pm$0.019&0.004$\pm$0.000&{0.048}$\pm$0.021\\
    \bottomrule
  \end{tabular}
  }
  \caption{\textbf{Generation results on the Enzymes and the Grid datasets.} Results of the baselines are taken from published papers \cite{jo2022score}. Hyphen (-) denotes that the results are not provided in the original paper. The best results are highlighted in bold (lower is better).  We report the MMD distance between the test datasets and generated graphs with the standard deviation.}
\label{Additional1}
\end{table*}

\begin{table*}[htbp]
  \centering
  \resizebox{\linewidth}{!}{
  \begin{tabular}{ccccccccc}
    \toprule
     Method & \makecell{Val. w/o corr. (\%) } $\uparrow$ & \makecell{NSPDK MMD} $\downarrow$ & FCD $\downarrow$&Validity $\uparrow$& Uniqueness $\uparrow$ & Novelty $\uparrow$\\
    \midrule
    GraphAF & 67&0.020±0.003&5.268±0.403&\textbf{100.00}&94.51&88.83 \\
    GraphDF & 82.67&0.063±0.001&10.816±0.020&\textbf{100.00}&97.62&\textbf{98.10}  \\
    MoFlow & 91.36±1.23&0.017±0.003&4.467±0.595&\textbf{100.00}±0.00&98.65±0.57&94.72±0.77   \\
    EDP-GNN & 47.52±3.60&0.005±0.001&2.680±0.221&\textbf{100.00}±0.00&\textbf{99.25}±0.05&86.58±1.85  \\
    GraphEBM & 8.22±2.24&0.030±0.004&6.143±0.411&\textbf{100.00}±0.00&97.90±0.14&97.01±0.17  \\
    $\text{GDSS}^*$& 95.79±1.93&0.003±0.000&2.813±0.278&\textbf{100.00}±0.00&98.02±0.63&82.55±3.11 \\
    \midrule
    HGDM (ours)&\textbf{98.04}±1.27&\textbf{0.002}±0.000&\textbf{2.13}±0.254&\textbf{100.00}±0.00& 97.27±0.71& 69.63±2.75 \\
    \bottomrule
  \end{tabular}
  }
  \caption{\textbf{Generation results on the QM9 dataset.} $*$ denotes that the results are obtained by running open-source codes. The results of HGDM are the means and the standard deviations of 3 runs. Other results of the baselines are taken from \cite{jo2022score}. The best results are highlighted in bold.}
  \label{Additional2}
\end{table*}

\begin{table*}[htbp]
  \centering
  \resizebox{\linewidth}{!}{
  \begin{tabular}{ccccccccc}
    \toprule
     Method & \makecell{Val. w/o corr. (\%) } $\uparrow$ & \makecell{NSPDK MMD} $\downarrow$ & FCD $\downarrow$&Validity $\uparrow$& Uniqueness $\uparrow$ & Novelty $\uparrow$\\
    \midrule
    GraphAF & 68&0.044±0.006&\textbf{16.289±0.482}&\textbf{100.00}&99.10&\textbf{100.00} \\
    GraphDF & 89.03&0.176±0.001&34.202±0.160&\textbf{100.00}&99.16&\textbf{100.00}  \\
    MoFlow & 63.11±5.17&0.046±0.002&20.931±0.184&\textbf{100.00}±0.00&\textbf{99.99}±0.01&\textbf{100.00}±0.00   \\
    EDP-GNN & 82.97±2.73&0.049±0.006&16.737±1.300&\textbf{100.00}±0.00&99.79±0.08&\textbf{100.00}±0.00  \\
    GraphEBM & 5.29±3.83&0.212±0.075&35.471±5.331&99.96±0.02&98.79±0.15&\textbf{100.00}±0.00  \\
    $\text{GDSS}^*$& \textbf{95.90±1.01}&0.019±0.001&16.621±1.213&\textbf{100.00}±0.00&99.67±0.14&\textbf{100.00}±0.00 \\
    \midrule
    HGDM (ours)&93.51±0.87&\textbf{0.016}±0.001&17.69±1.146&\textbf{100.00}±0.00& 99.82±0.18 & \textbf{100.00}±0.00\\
    \bottomrule
  \end{tabular}
  }
  \caption{\textbf{Generation results on the ZINC250k dataset.}  $*$ denotes that the results are obtained by running open-source codes. The results of HGDM are the means and the standard deviations of 3 runs. Other results of the baselines are taken from \cite{jo2022score}. The best results are highlighted in bold.}
  \label{Additional3}
\end{table*}
\begin{table*}[htbp]
  
  \centering
  \resizebox{\linewidth}{!}{
  \begin{tabular}{cccccccc}
    \toprule
    
    &Hyperparameter&Ego-small&Community-small&Enzymes&Grid&QM9&ZINC250k \\
    \midrule
    \multirow{2}{*}{$f_{enc}$}&Number of HGAT layers&3&3&3&3&3&3 \\
    &Hidden dimension&32&32&32&32&16&16\\
    \midrule
    \multirow{2}{*}{$f_{dec}$}&Number of HGAT layers&0&0&0&3&0&3 \\
    &Hidden dimension&32&32&32&32&16&16\\
    \midrule
    \multirow{2}{*}{$s_\theta$}&Number of HGAT layers&2&3&5&5&2&2 \\
    &Hidden dimension&32&32&32&32&16&16\\
    \midrule
    \multirow{4}{*}{$s_\phi$}&Number of attention heads&4&4&4&4&4&4\\
    &Number of initial channels&2&2&2&2&2&2\\
    &Number of hidden channels&8&8&8&8&8&8\\
    &Number of final channels&4&4&4&4&4&4\\
    &Number of GCN layers&5&5&7&7&3&6\\
    &Hidden dimension&32&32&32&32&16&16\\
    \midrule
    \multirow{4}{*}{SDE for X}&Type &VE &VP &VP &VP &VP &VP \\
    &Number of sampling steps &1000 &1000 &1000 &1000 &1000 &1000 \\
    &$\beta_{min}$ &0.1 &0.1 &0.1 &0.1 &0.1 &0.1 \\
    &$\beta_{max}$ &4.0 &1.0 &1.0 &7.0 &2.0 &1.0 \\
    \midrule
    \multirow{4}{*}{SDE for A}&Type &VP &VP &VE &VP &VE &VE \\
    &Number of sampling steps &1000 &1000 &1000 &1000 &1000 &1000 \\
    &$\beta_{min}$ &0.1 &0.1 &0.2 &0.2 &0.1 &0.2 \\
    &$\beta_{max}$ &1.0 &1.0 &1.0 &0.8 &1.0 &1.0\\
    \midrule
    \multirow{3}{*}{Solver}&Type &EM + Langevin &Rev. + Langevin &Rev. + Langevin& Rev. + Langevin &EM + Langevin &Rev. + Langevin \\
    &SNR& 0.25 &0.05 &0.5 &0.25 &0.15&0.2 \\
    &Scale coefficient&1.0& 0.9 &0.9 &0.9 &0.8 &0.4\\
    \midrule
    \multirow{7}{*}{Train}&Optimizer &Riemannian Adam &Riemannian Adam &Riemannian Adam &Riemannian Adam &Riemannian Adam &Riemannian Adam \\
    &kl weight for VAE&$1 \times 10^{-5}$&$1 \times 10^{-2}$&$1 \times 10^{-5}$&$1 \times 10^{-2}$& $1 \times 10^{-2}$&$1 \times 10^{-2}$\\
    &Learning rate &$1 \times 10^{-2}$ &$1 \times 10^{-2}$ &$1 \times 10^{-2}$ &$1 \times 10^{-2}$ &$5 \times 10^{-3}$ &$5 \times 10^{-3}$ \\
    &Weight decay &$1 \times 10^{-4}$ &$1 \times 10^{-4}$ &$1 \times 10^{-4}$ &$1 \times 10^{-4}$ &$1 \times 10^{-4}$ &$1 \times 10^{-4}$ \\
    &Batch size &128 &128 &64 &8 &1024 &1024 \\
    &Number of epochs &5000 &5000 &5000 &5000 &300 &500\\
    &EMA&-&-&0.999&0.999&-&-\\
    \bottomrule
  \end{tabular}
  }
  \caption{\textbf{Hyperparameters of HGDM} used in the generic graph generation tasks and the molecule generation tasks. We provide the hyperparameters of the HVAE, the score-based models ($s_\theta$ and $s_\phi$), the diffusion processes (SDE for X and A), the SDE solver, and the training.}
  \label{Hyperparameters}
\end{table*}

\section{Experimental Details}
In this section, we explain the details of the experiments including the generic graph generation tasks, and the molecule generation tasks. We describe the implementation details of HGDM and further provide the hyperparameters used in the experiments in Table \ref{Hyperparameters}. For a fair comparison, we keep most of the hyperparameters consistent with GDSS \cite{jo2022score}.

\subsection{Generic Graph Generation}
For a fair evaluation of the generic graph generation task, we follow the standard setting of existing works \cite{jo2022score,you2018graphrnn,liu2019graph,niu2020permutation} from the node features to the data splitting. We initialize the node features as the one-hot encoding of the degrees. We perform the grid search to choose the best signal-to-noise ratio (SNR) in \{0.05, 0.1, 0.15, 0.2, 0.25, 0.3, 0.35, 0.4, 0.45, 0.5\} and the scale coefficient in the \{0.1, 0.2, 0.3, 0.4, 0.5, 0.6, 0.7, 0.8, 0.9, 1.0\}. We select the best results with the lowest average of degree, clustering coefficient, and orbit. After generating the samples by simulating the reverse diffusion process, we quantize the entries of the adjacency matrices with the operator $1_{x>0.5}$ to obtain the 0-1 adjacency matrix. 
We fix the curvature of the hyperbolic space to $-0.01$, which empirically improves performance. We optimize HGDM with Riemannian Adam \cite{becigneul2018riemannian,kochurov2020geoopt}. 

We report the remaining hyperparameters used in the experiment in Table \ref{Hyperparameters}. The decoder in Table \ref{Hyperparameters} with 0 layers of HGAT consists of only one centroid-distance layer \cite{liu2019hyperbolic} and one linear layer. The weight of edge loss $\mathcal{L}_{\textit{edge}}$ is set to 0.01 for all datasets.

\subsection{Molecule Generation}
Following \citet{jo2022score}, each molecule is preprocessed into a graph with the node features $X \in \{0, 1\}^{N\times F}$ and the adjacency matrix $A \in \{0, 1, 2, 3\}^{N\times N}$, where $N$ is the maximum number of atoms in a molecule of the dataset, and $F$ is the number of atom types. The entries of A indicate the bond types. We kekulize the molecules using the RDKit library \cite{landrum2016rdkit} and remove all hydrogen atoms following the standard procedure \cite{shi2020graphaf,luo2021graphdf}. We make use of the valency correction proposed by \citet{zang2020moflow}. We perform the grid search to choose the best signal-to-noise ratio (SNR) in \{0.05, 0.1, 0.15, 0.2\} and the scale coefficient in \{0.1, 0.2, 0.3, 0.4, 0.5, 0.6, 0.7, 0.8, 0.9, 1.0\}. We choose the hyperparameters that exhibit the best FCD value. After generating the samples by simulating the reverse diffusion process, we quantize the entries of the adjacency matrices to $\{0, 1, 2, 3\}$ by clipping the values as: $(-\infty, 0.5)$ to 0, the values of $[0.5, 1.5)$ to 1, the values of $[1.5, 2.5)$ to 2, and the values of $[2.5, +\infty)$ to 3. Then We use a pre-trained decoder from HVAE to recover the atom type. We fix the curvature of the hyperbolic space to $-0.01$. We optimize our model with Riemannian Adam \cite{becigneul2018riemannian,kochurov2020geoopt}. 

We report the remaining hyperparameters used in the experiment in Table \ref{Hyperparameters}. The decoder in Table \ref{Hyperparameters} with 0 layers of HGAT consists of only one centroid-distance layer \cite{liu2019hyperbolic} and one linear layer. The weight of edge loss $\mathcal{L}_{\textit{edge}}$ is set to 0.01 for all datasets.

\subsection{Computing Resources}
For all the experiments, we utilize PyTorch \cite{paszke2019pytorch} to implement HGDM and train the score models on GeForce RTX 4090 GPU. For the molecule generation tasks, the inference time of HGDM and GDSS is measured on 1 GeForce RTX 4090 GPU and 24 CPU cores.

\section{Visualization}
In this section, we provide the visualizations of the graphs generated by our HGDM for the generic graph generation tasks and molecule generation tasks.

\subsection{Molecule Examples}
\begin{figure*}[tbp]
\centering
    \includegraphics[width=1.8\columnwidth]{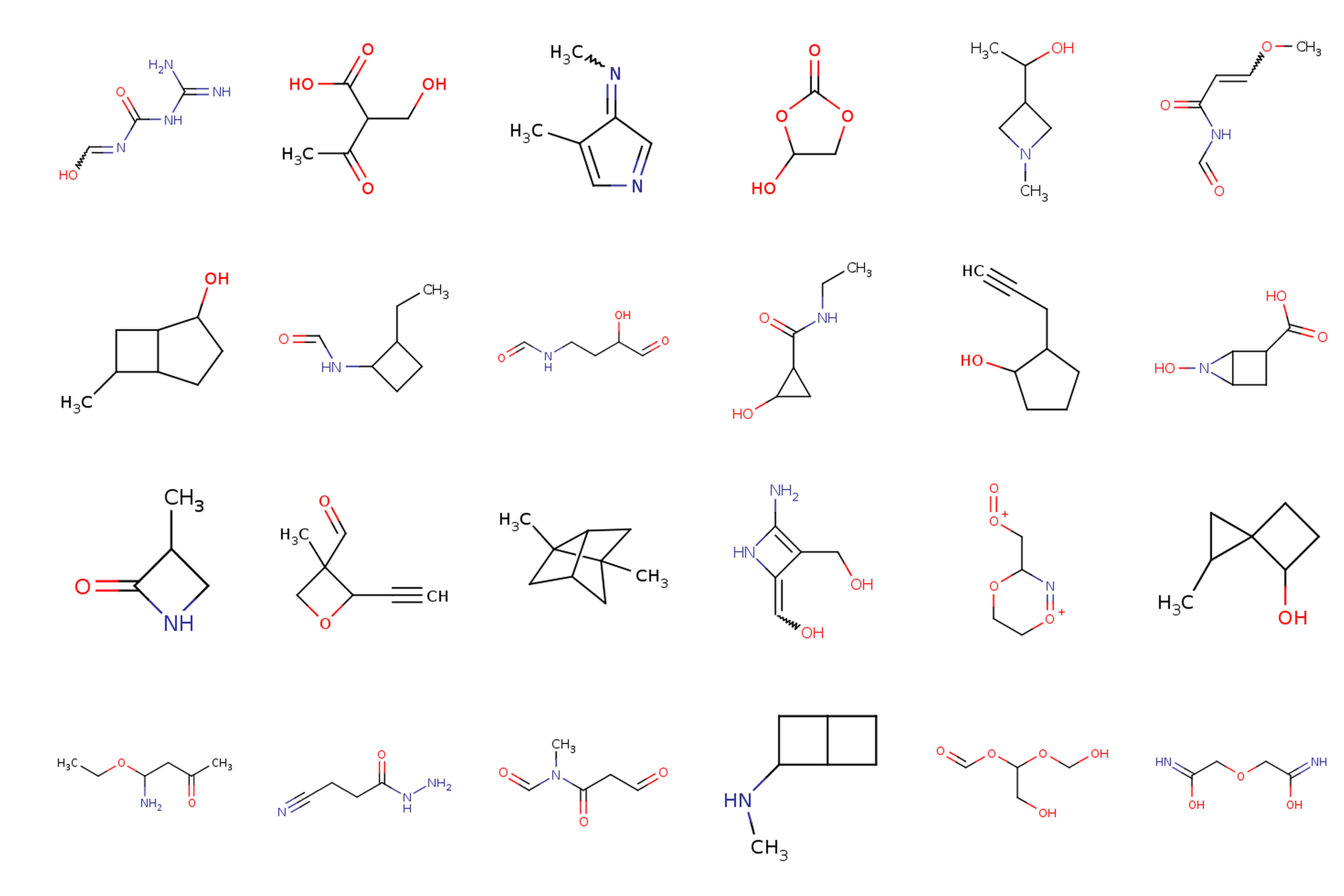}
    \caption{Random samples taken from the HGDM trained on QM9.}
    \label{QM9}
\end{figure*}

\begin{figure*}[tbp]
\centering
    \includegraphics[width=1.8\columnwidth]{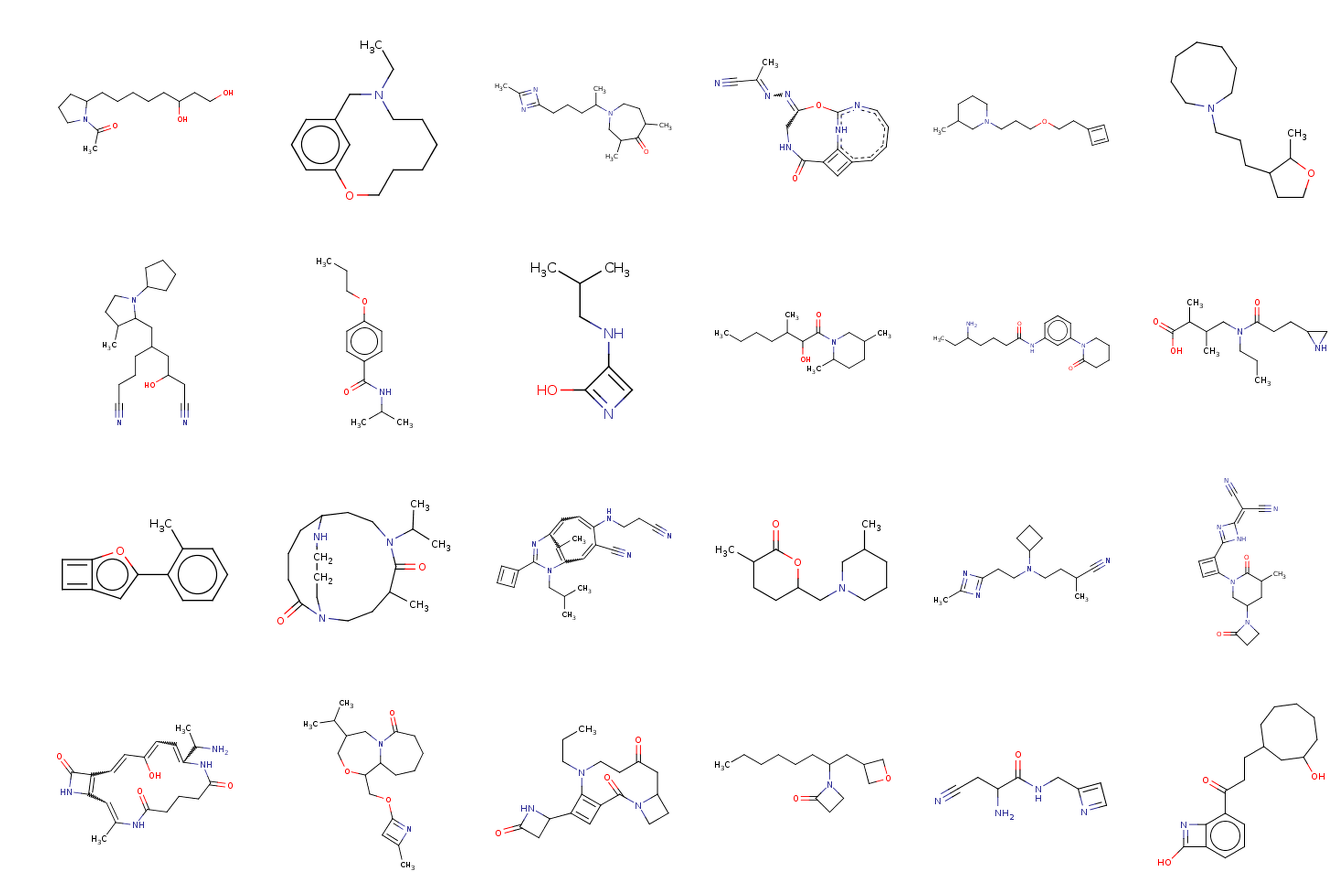}
    \caption{Random samples taken from the HGDM trained on ZINC250k.}
    \label{ZINC}
\end{figure*}

We visualize a subset of molecules generated by HGDM in Figure \ref{QM9} and \ref{ZINC}.

\subsection{Generic Graph Examples}
We visualize the graphs from the training datasets and the generated graphs of HGDM for each dataset in Figure 5-8. The visualized graphs are the randomly selected samples from the training datasets and the generated graph set. We additionally provide the information of the number of edges $e$ and the number of nodes $n$ of each graph.

\begin{figure*}[htbp]
\centering
\subfigure[Training Data]{
        \includegraphics[width=0.45\textwidth]{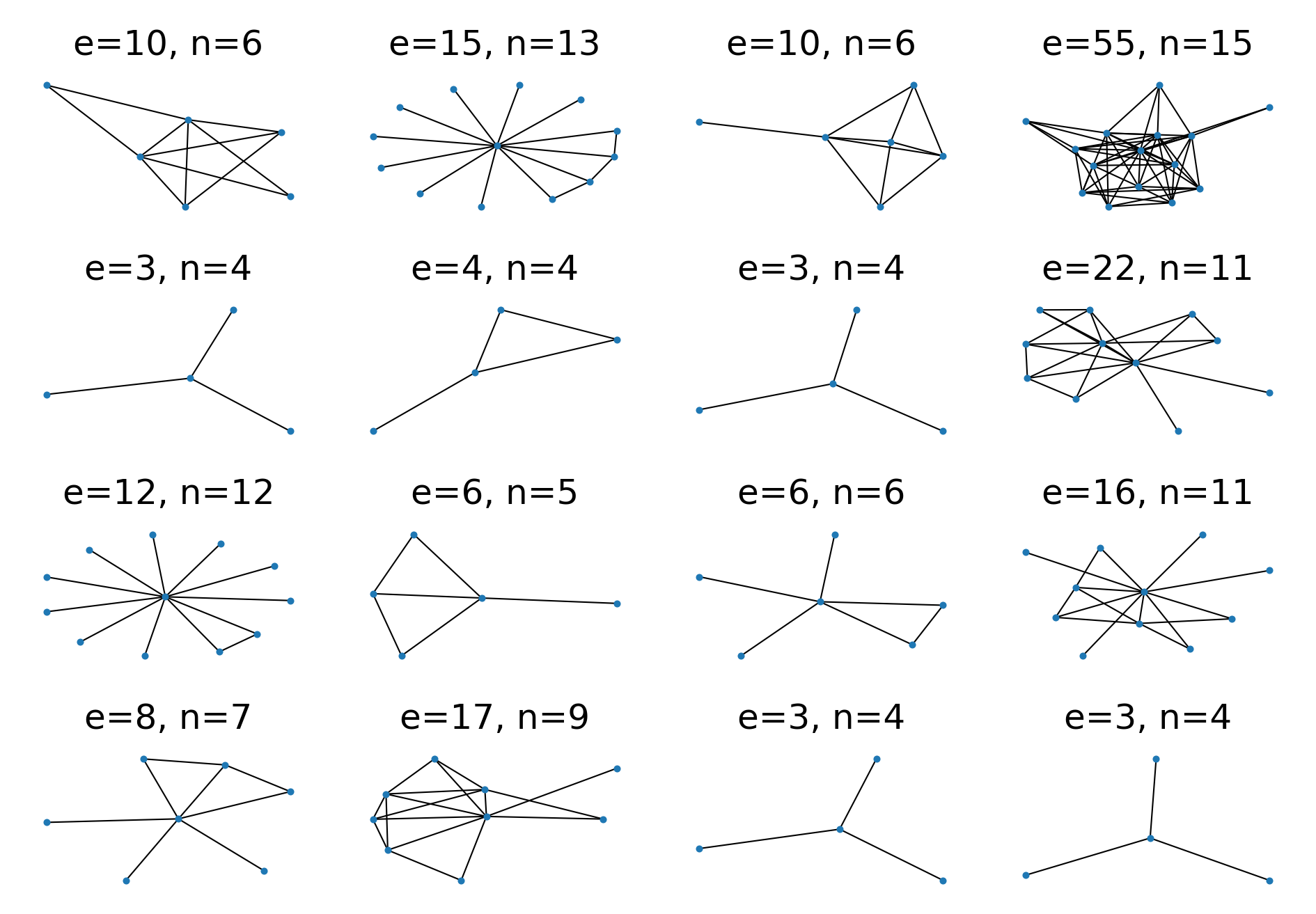}
    } 
    \hspace{0.3in}
    \subfigure[HGDM (ours)]{
        \includegraphics[width=0.45\textwidth]{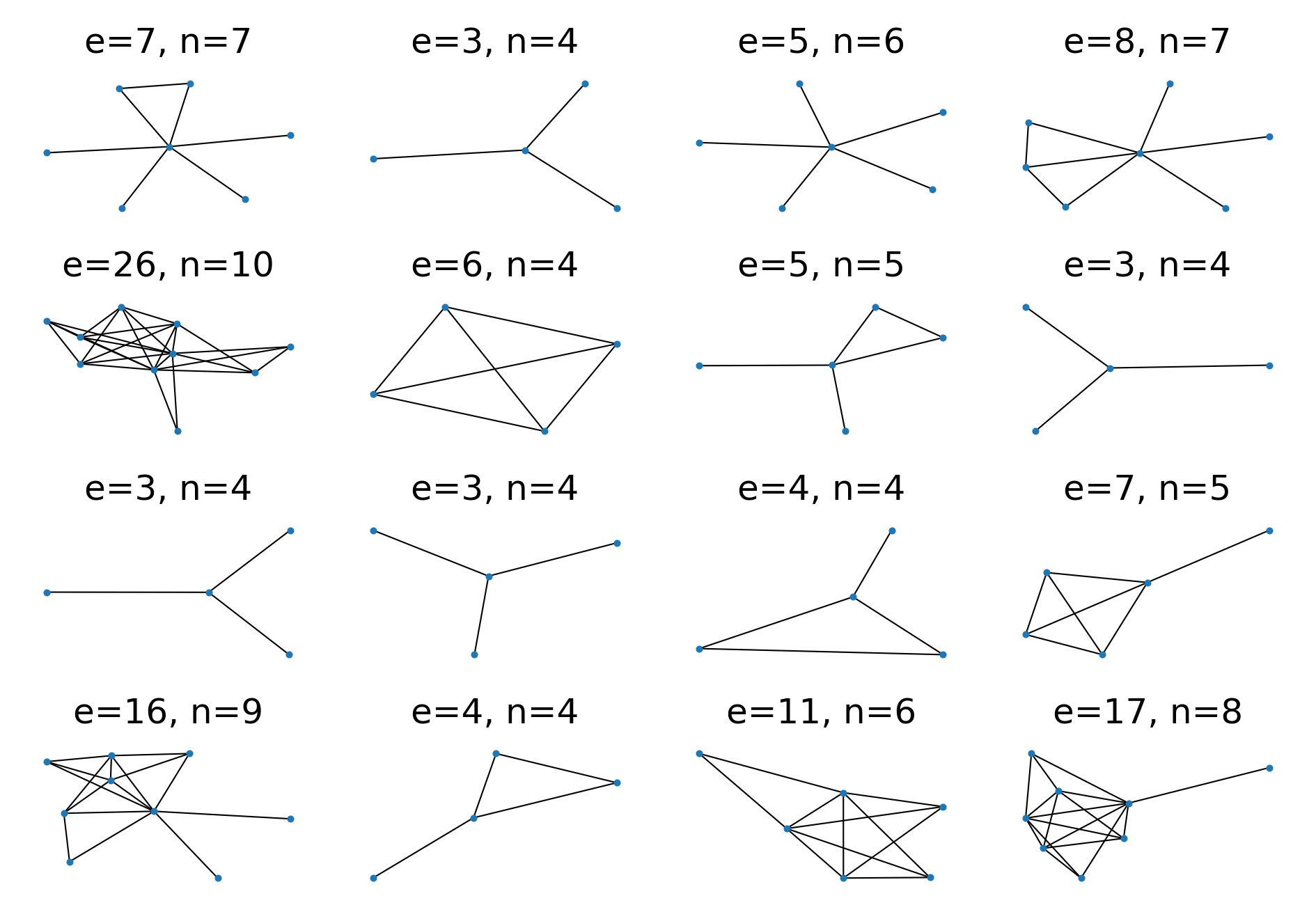}
    }
    \caption{Visualization of the graphs from the Ego small dataset and the generated graphs of HGDM.}
    \label{fig:vis_es}
\end{figure*}
\begin{figure*}[htbp]
\centering
\subfigure[Training Data]{
        \includegraphics[width=0.45\textwidth]{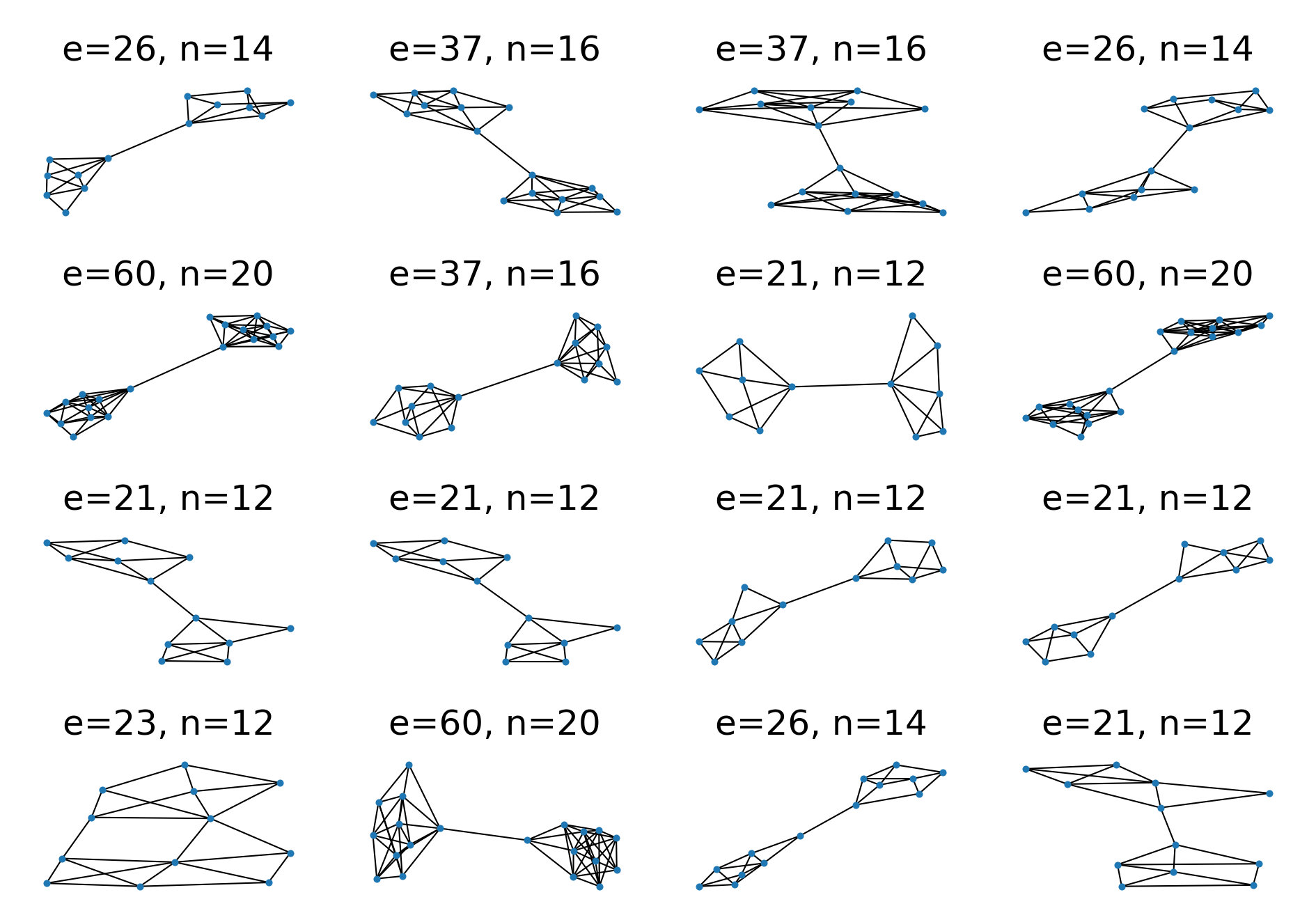}
    } 
    \hspace{0.3in}
    \subfigure[HGDM (ours)]{
        \includegraphics[width=0.45\textwidth]{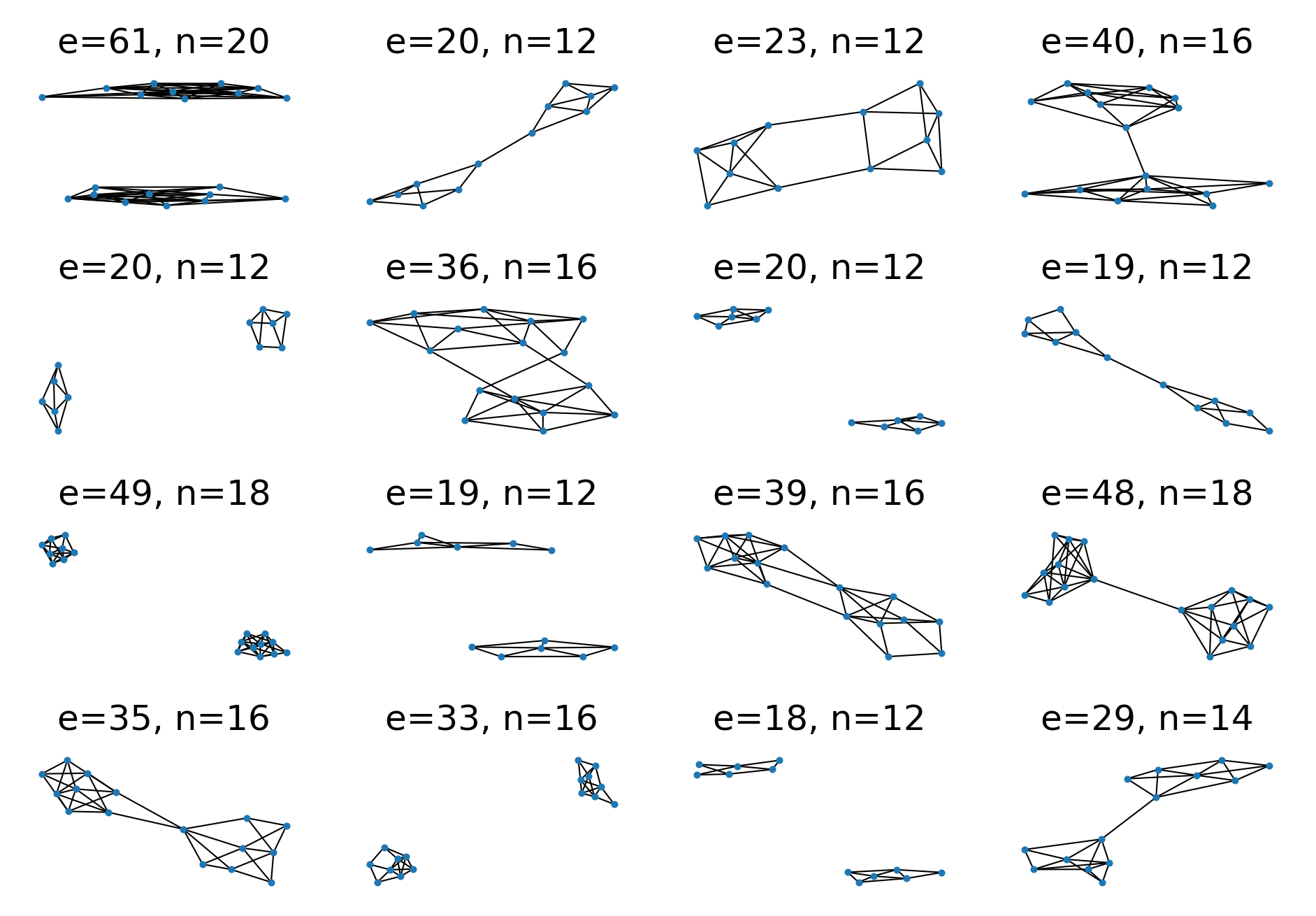}
    }
    \caption{Visualization of the graphs from the Community small dataset and the generated graphs of HGDM.}
    \label{fig:vis_cs}
\end{figure*}
\begin{figure*}[htbp]
\centering
\subfigure[Training Data]{
        \includegraphics[width=0.45\textwidth]{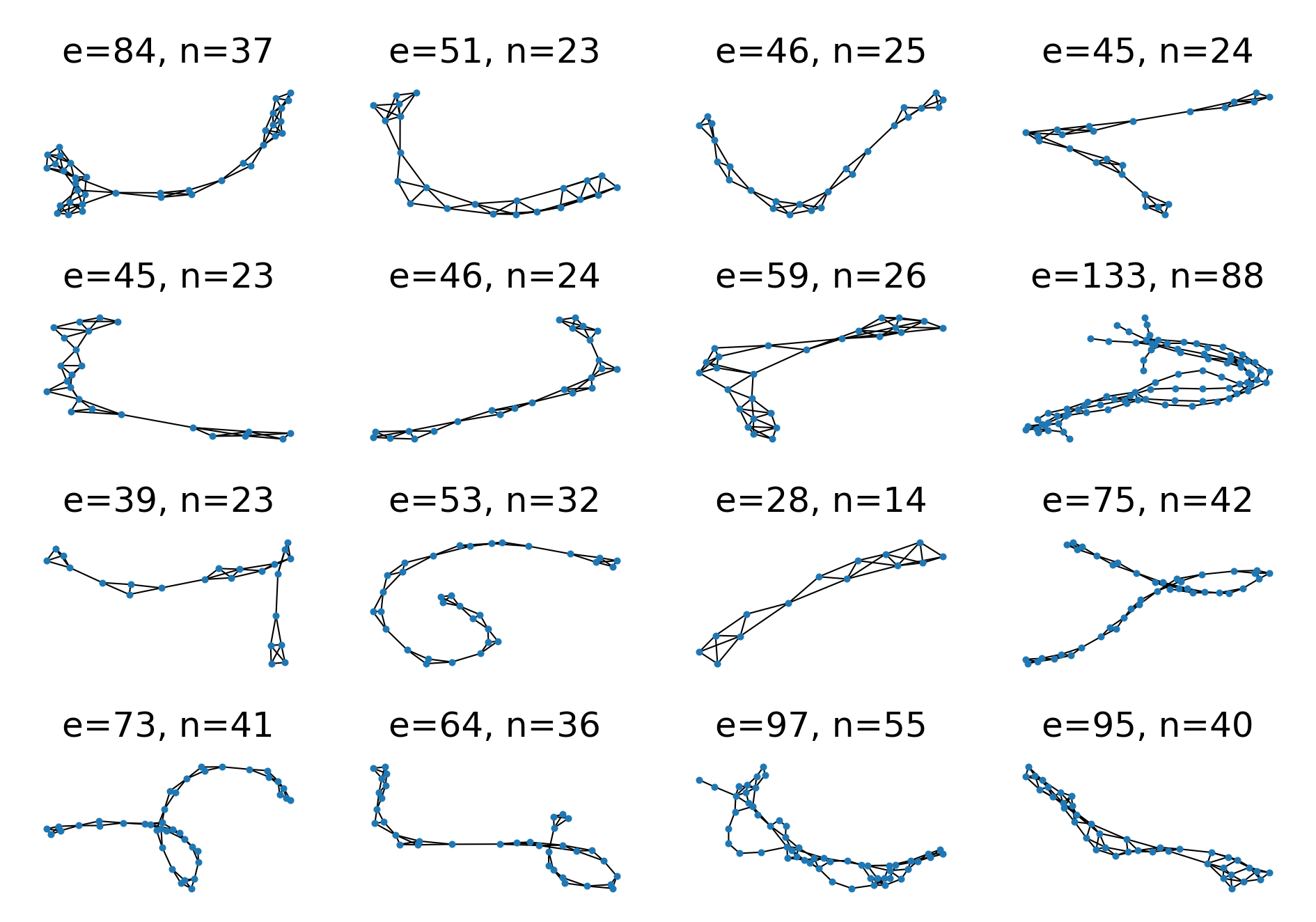}
    } 
    \hspace{0.3in}
    \subfigure[HGDM (ours)]{
        \includegraphics[width=0.45\textwidth]{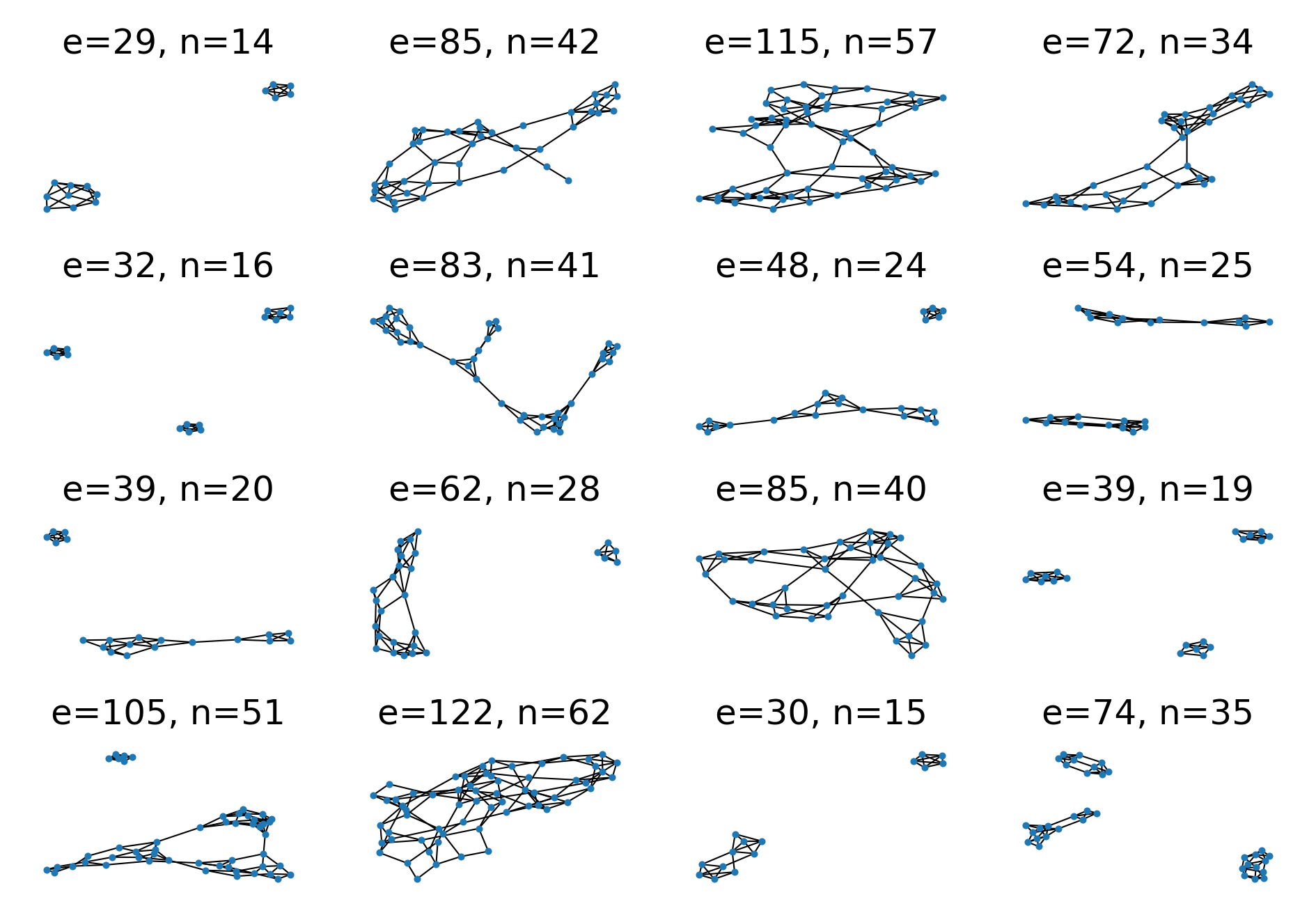}
    }
    \caption{Visualization of the graphs from the Enzymes dataset and the generated graphs of HGDM.}
    \label{fig:vis_en}
\end{figure*}
\begin{figure*}[htbp]
\centering
\subfigure[Training Data]{
        \includegraphics[width=0.45\textwidth]{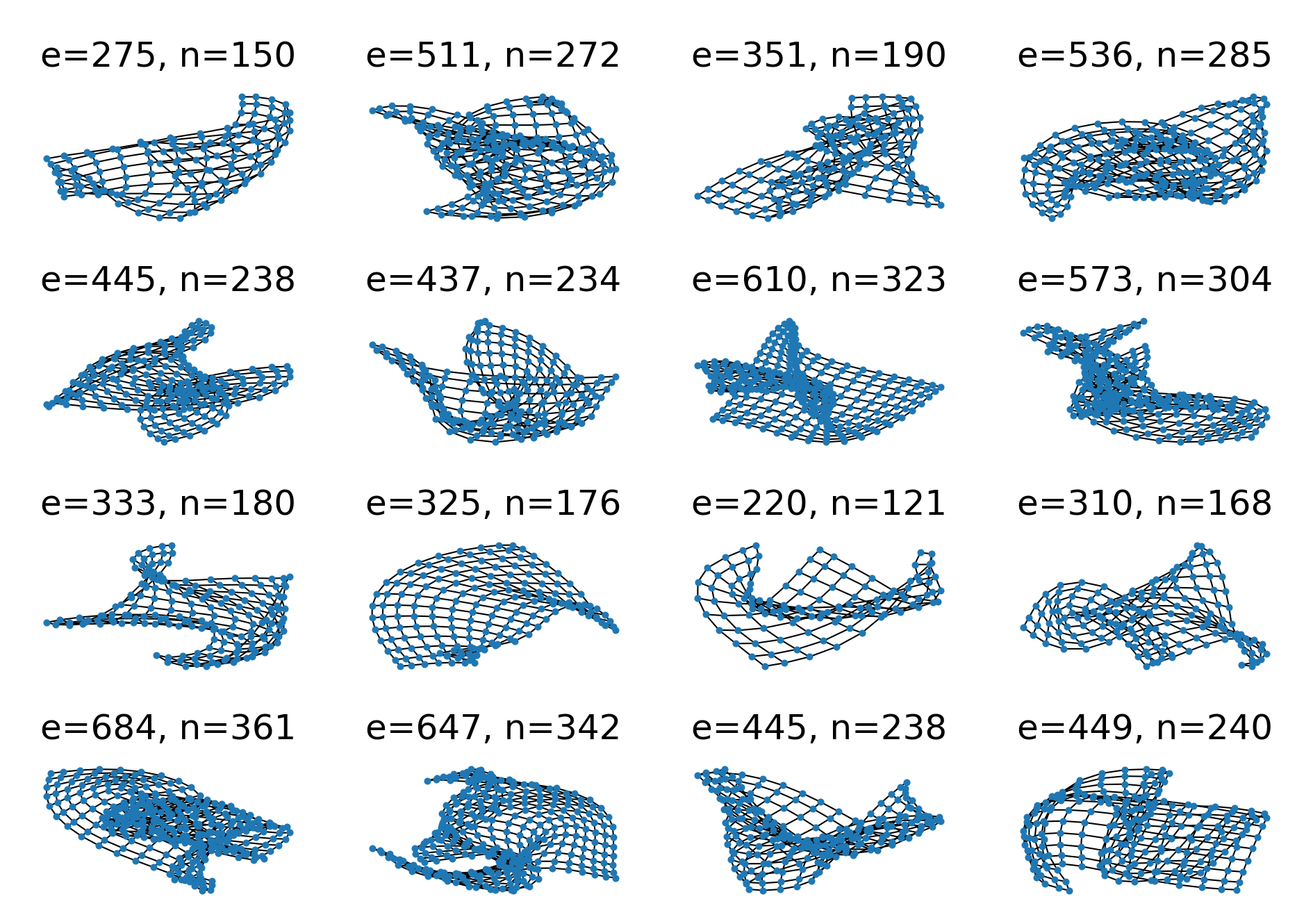}
    } 
    \hspace{0.3in}
    \subfigure[HGDM (ours)]{
        \includegraphics[width=0.45\textwidth]{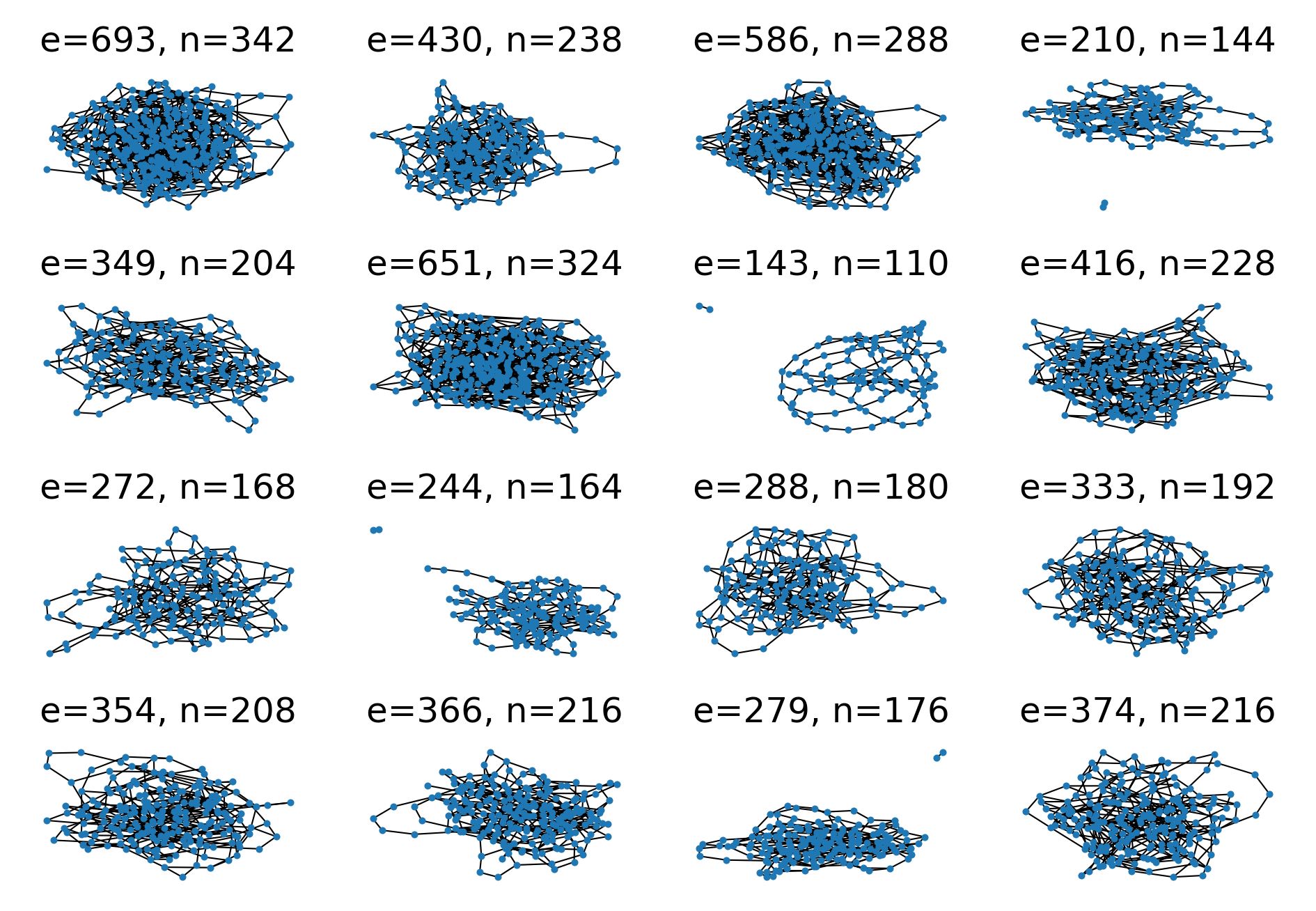}
    }
    \caption{Visualization of the graphs from the Grid dataset and the generated graphs of HGDM.}
    \label{fig:vis_grid}
\end{figure*}

\bigskip

\end{document}